\pdfoutput=1

\documentclass[11pt]{article}

\usepackage{acl}

\usepackage{times}
\usepackage{latexsym}
\usepackage[most]{tcolorbox}

\usepackage[T1]{fontenc}

\usepackage[utf8]{inputenc}

\usepackage{microtype}

\usepackage{inconsolata}

\usepackage{hyperref}
\usepackage{url}
\usepackage{booktabs}
\usepackage{multirow}
\usepackage{multicol}
\usepackage{makecell}
\usepackage{graphicx}
\usepackage{caption}
\usepackage{tabularx}
\usepackage{nicematrix}
\usepackage{threeparttable}

\title{TRAM: Benchmarking Temporal Reasoning for Large Language Models}

\author{Yuqing Wang \\
  Stanford University \\
  \texttt{ywang216@stanford.edu} \\\And
  Yun Zhao \\
  Meta Platforms, Inc. \\
  \texttt{yunzhao20@meta.com} \\}

\begin{document}
\maketitle
\begin{abstract}
Reasoning about time is essential for understanding the nuances of events described in natural language. Previous research on this topic has been limited in scope, characterized by a lack of standardized benchmarks that would allow for consistent evaluations across different studies. In this paper, we introduce TRAM, a temporal reasoning benchmark composed of ten datasets, encompassing various temporal aspects of events such as order, arithmetic, frequency, and duration, designed to facilitate a comprehensive evaluation of the TeR capabilities of large language models (LLMs). We evaluate popular LLMs like GPT-4 and Llama2 in zero-shot and few-shot scenarios, and establish baselines with BERT-based and domain-specific models. Our findings indicate that the best-performing model lags significantly behind human performance. It is our aspiration that TRAM will spur further progress in enhancing the TeR capabilities of LLMs. Our
data and code are available at \url{https://github.com/EternityYW/TRAM-Benchmark}.
\end{abstract}

\section{Introduction}
Temporal reasoning is fundamental for humans to understand the world and distinguish between everyday events. For instance, when given the activities “watching a movie” and “watching a sunset”, we intuitively recognize that, though both are time-bound, a movie typically lasts longer than a sunset. Moreover, while movies can be watched repeatedly, sunsets transpire once a day. Such innate comprehension is not just about sequencing events or understanding durations; it extends to more intricate aspects of time, allowing us to make sense of complex narratives and the causality of events. Despite advancements in natural language processing (NLP) and the advent of large language models~\citep{min2021recent, zhao2023survey, wang2023large}, mastering temporal reasoning remains a significant challenge due to its intricate nature, the variability of temporal expressions, and the need for contextual understanding.

Recent work in temporal reasoning (TeR) has primarily focused on time-sensitive question-answering~\citep{zhou2019going, chen2021dataset, dhingra2022time, tan-etal-2023-towards}, demonstrating that despite significant advancements in NLP, current language models have yet to reach human-level performance in this domain. Furthermore, these studies, while addressing explicit temporal elements such as order, duration, and time-event relations, overlook more complex aspects of TeR, like temporal narratives and causality. Importantly, the establishment of a unified framework including broad facets of TeR has not yet been achieved.

\begin{figure}[ht]
\centering
\includegraphics[width=.49\textwidth]{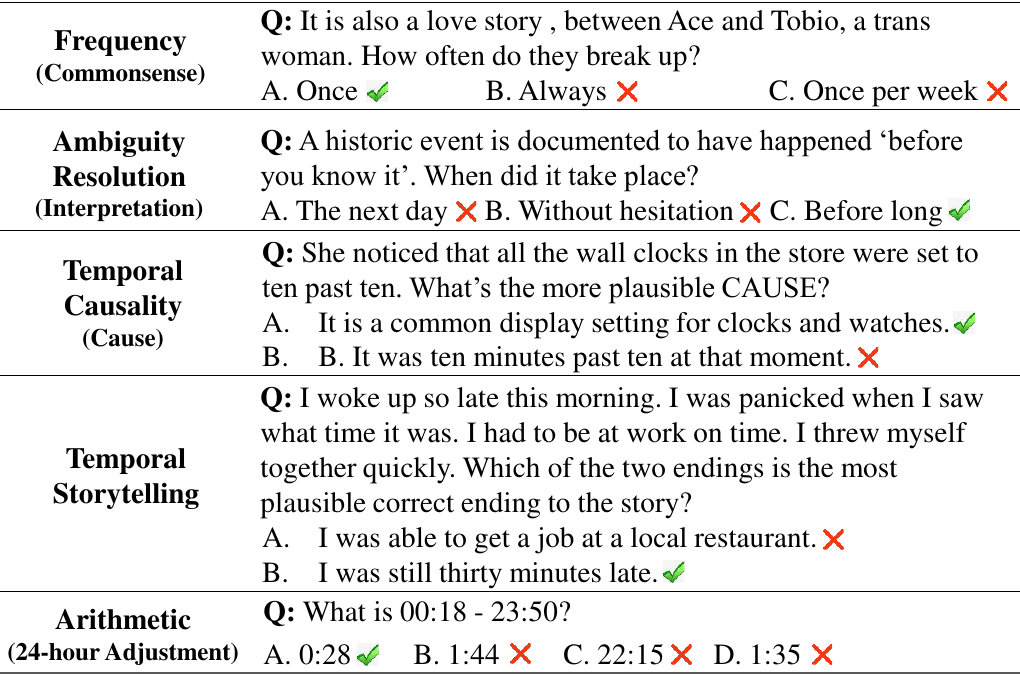}  
\caption{Example questions in TRAM.}
\label{fig: data_examples}
\end{figure}

To facilitate research in this direction, we present the \textbf{T}emporal \textbf{R}easoning for large l\textbf{A}nguage \textbf{M}odel benchmark (or TRAM for short), a collection of ten temporal reasoning tasks. These tasks range from foundational understanding (e.g., duration, frequency) to advanced temporal interpretations and computations (e.g., arithmetic, causality). Each task consists of one or more subtasks, all of which are specifically crafted to assess a model's TeR capabilities across varying levels of understanding and difficulty. In total, our benchmark includes 38 distinct subtasks, comprising a total of 526.7k questions. Answers have been derived through a combination of expert annotations and programmatic generation. Diverging from prior TeR research and in line with~\citep{hendrycks2020measuring}, our questions are formatted as straightforward multiple-choice tests rather than generative tasks, thereby more appropriately evaluating LLMs. Example questions in TRAM are shown in Figure~\ref{fig: data_examples}.

To gain deeper insight into the TeR challenges posed by TRAM, we extensively evaluate several prominent language models, including BERT~\citep{kenton2019bert}, RoBERTa~\citep{liu2019roberta}, the domain-specific TeR model RST~\citep{yuan2022restructured}, and recent LLMs including Llama2~\citep{touvron2023llama}, Gemini Pro~\citep{team2023gemini}, GPT-3.5, and GPT-4~\citep{openai2023gpt4}. We use limited training data to fine-tune BERT-style and RST models. LLMs are evaluated using standard and chain-of-thought prompting under zero-shot and few-shot learning paradigms. Our results indicate that GPT-4 excels in most tasks, achieving an average accuracy of up to 84.4\%. Moreover, we observe notable performance disparities across tasks among the models. Despite the impressive performance of GPT-4, it falls short of human proficiency by over 10\%, highlighting significant room for LLMs to improve their TeR capabilities. Manual error analysis shows that models struggle with nuanced understanding and interpreting implicit cues across all task categories.

In summary, our contributions are threefold:
\begin{enumerate}
\item[(1)] We introduce TRAM, a comprehensive collection of ten distinct TeR tasks featuring 526.7k questions presented in a multiple-choice format. Ranging from foundational temporal concepts to intricate temporal interpretations, TRAM serves as a unified framework for assessing the TeR capabilities of LLMs.
\item[(2)] We conduct extensive experiments on TRAM, evaluating leading language models including BERT-style models, a TeR-specific model, and LLMs such as Llama2 and GPT-4. Our results reveal that even the best-performing model notably falls short of human-level performance, underscoring the opportunities for continued research in this area.
\item[(3)] Manual error analysis reveals consistent TeR challenges for current LLMs, particularly in nuanced comprehension and decoding implicit temporal cues, highlighting the need for further research to enhance LLM capabilities in addressing these specific errors.
\end{enumerate}

\section{Related Work}
Our proposal for a comprehensive TeR benchmark builds on the evolution of datasets in this domain while addressing the lack of a unified system for evaluation. The modern NLP landscape sets the stage for a nuanced evaluation of both pretrained models and LLM paradigms.

\noindent \textbf{Temporal Reasoning Benchmarks} In the realm of TeR, several datasets have emerged to address distinct challenges. Early benchmarks, such as TimeBank~\citep{pustejovsky2003timebank}, predominantly focused on temporal relations. TempEval-3~\citep{uzzaman2013semeval} broadened the scope by introducing multiple tasks, including temporal entity extraction and temporal relation extraction. Recently, there has been a surge in the development of time-sensitive question-answering datasets like MCTACO~\citep{zhou2019going}, Time-sensitive QA~\citep{chen2021dataset}, TEMPLAMA~\citep{dhingra2022time}, TEMPREASON~\citep{tan-etal-2023-towards}, and MenatQA~\citep{wei2023menatqa}. However, these datasets often specialize in narrower aspects of TeR, such as duration, frequency, or event-time relations. In contrast, our benchmark offers a comprehensive scope of TeR, addressing diverse levels and dimensions of understanding about time, aiming to provide a more complete representation of TeR challenges than previously available datasets. 

\noindent \textbf{Training Paradigms in LLMs} 
In NLP research, pretraining language models on vast amounts of
diverse texts has become standard practice. Through this process, the models encapsulate a broad spectrum of information across various domains. BERT-based models like BERT~\citep{kenton2019bert} and RoBERTa~\citep{liu2019roberta} are representative examples. These models have been applied to a diverse set of tasks, including disease prediction~\citep{zhao2021empirical}, text classification~\citep{wang2022integrating}, time series analysis~\citep{wang2022enhancing}, and more. However, the advent of GPT-3~\citep{brown2020language} shifted the focus towards minimal fine-tuning approaches, such as zero-shot and few-shot learning, allowing models to adapt to new tasks with only a few training examples~\citep{brown2020language}. This transition has spurred the development of advanced prompting techniques aimed at enhancing the understanding and reasoning capabilities of LLMs. Some representative prompting methods include CoT prompting~\citep{wei2022chain}, self-consistency~\citep{wang2022self}, tree-of-thought prompting~\citep{yao2023tree}, and metacognitive prompting~\citep{wang2023metacognitive}. In this work, we establish baseline evaluations by considering traditional BERT-based models, a domain-specific TeR model, and recent LLMs such as Llama2 and GPT-4 to provide a comprehensive understanding of their strengths and limitations in diverse TeR tasks.

\section{Tasks and Datasets}
TRAM encompasses ten distinct tasks, presented as multiple-choice questions (MCQs) across a range of time-related domains. For clarity and directness, we ensure that each question has only one correct answer. The main purpose of TRAM is to spur further research into the advanced TeR capabilities of LLMs. Overall, these tasks fall under three distinct groups.
(1) \textit{Foundational Temporal Understanding Tasks:} Covering basic temporal comprehension, this group incorporates tasks such as ordering, frequency, duration, and typical time.
(2) \textit{Temporal Interpretation and Computation Tasks:} Centered on the interpretative and computational aspects of time, this group includes tasks like ambiguity resolution and arithmetic.
(3) \textit{Advanced Temporal and Conceptual Understanding Tasks:} Dedicated to exploring intricate temporal relationships and narratives, this category features tasks like relation, temporal NLI, causality, and storytelling. In this work, certain task names, such as ‘relation' and ‘causality', can have varied interpretations across different contexts. However, they are specifically emphasized for their temporal aspects in this work. Although we might occasionally omit the term ‘temporal' for brevity, readers should note that the tasks are centered on time-related elements.

In TRAM, each task is designed with one or more problem types, ensuring diverse representation across data attributes, complexities, and domains. The benchmark includes 526,668 problems in total. For each dataset, we introduce a few-shot development set, with five questions per category, and a separate test set for evaluation. Table~\ref{tab: dataset} provides an overview of these tasks, and more details can be found in Appendix~\ref{data}. The majority of tasks employ accuracy as the evaluation metric due to their straightforward MCQ structure. For tasks like ‘relation' and ‘temporal NLI', which exhibit an imbalanced label distribution inherent to their fixed class structure, both accuracy and the F1 score are utilized, even when they are presented as MCQs.

\begin{table*}[h]
\centering 
\caption{Overview of ten tasks included in TRAM. The “Data Size” column aggregates totals from both the development and test sets.  “\textit{K}-Way MC” signifies a multiple-choice response format with \textit{K} options. \textit{Amb. Res.} denotes Ambiguity Resolution. \textit{NLI} stands for natural language inference. “Same” indicates the text source is the same as the row above.}
\resizebox{\linewidth}{!}{%
\vspace{1cm}
\begin{tabular}{cccccc}
    \toprule
    \textbf{Task} & \textbf{Data Size} & \textbf{$\#$ Problem Types} & \textbf{Metrics} & \textbf{Answer Type} & \textbf{Text Sources} \\ 
    \midrule
    \multicolumn{6}{c}{Foundational Temporal Understanding Tasks} \\
    \midrule
    Ordering & 29,462 & 2 & Acc. &  3-Way MC & MCTACO$^{1}$, Wikipedia, Misc.\\
    Frequency & 4,658 & 6 & Acc. & 3-Way MC & MCTACO$^{1}$, SQuAD$^{2}$, Misc. \\
    Duration & 7,232 & 7 & Acc. & 3-Way MC & Same\\
    Typical Time & 13,018 & 4 & Acc.& 3-Way MC & Same \\
    \midrule
    \multicolumn{6}{c}{Temporal Interpretation and Computation Tasks} \\
    \midrule
    Amb. Res. & 3,649 & 5 & Acc. & 3-Way MC & Misc. \\
    Arithmetic & 15,629 & 9 & Acc. & 4-Way MC & Same \\
    \midrule
    \multicolumn{6}{c}{Advanced Temporal and Conceptual Understanding Tasks} \\
    \midrule
    Relation & 102,462 & 1 & Acc./F1 & 3-Way MC & TempEval-3$^{3}$\\
    Temporal NLI & 282,144 & 1 & Acc./F1 & 3-Way MC & MNLI$^{4}$, SNLI$^{5}$\\
    Causality & 1,200 & 2 & Acc. & 2-Way MC & COPA$^{6}$, Misc.\\
    Storytelling & 67,214 & 1 & Acc. & 2-Way MC & ROC$^{7}$, SCT$^{8}$ \\
    \bottomrule
\end{tabular}}
\footnotesize 
$^{1}$~\citep{zhou2019going}, $^{2}$~\citep{rajpurkar2016squad},
$^{3}$~\citep{uzzaman2013semeval}, \\ $^{4}$~\citep{williams2018broad}, 
$^{5}$~\citep{bowman2015large}, $^{6}$~\citep{roemmele2011choice}, \\
$^{7}$~\citep{mostafazadeh2016corpus}, $^{8}$~\citep{mostafazadeh2017lsdsem}
\label{tab: dataset}
\end{table*}

\subsection{Foundational Temporal Understanding Tasks}
This group of tasks is fundamental for assessing a model's proficiency in core temporal concepts. For the tasks below, data from the Multiple
Choice TemporAl COmmon-sense (MCTACO) dataset incorporates both validation and test sets, while data from the Stanford Question Answering Dataset (SQuAD) dataset includes both training and validation sets. Unless otherwise mentioned, the options for each dataset are generated through a blend of human curation and algorithmic processes, tailored to each specific task. For instance, in the ordering task, correct answers strictly adhere to the accurate chronological sequence of events, while incorrect choices are formed through random permutations. 

\noindent \textbf{Ordering} The temporal ordering task evaluates a model's ability to understand the sequence in which events occur. This task is divided into two primary problem types. For \textit{commonsense} problems, we mainly source questions from the MCTACO dataset~\citep{zhou2019going}, specifically targeting subcategories related to temporal ordering. For each individual question selected from this dataset, we pose questions in the format, “Is \{candidate answer\} possible?” While MCTACO's expected answers are “true” or “false”, we introduce another layer of complexity by also including an “undetermined” option. Additionally, we curate another set of commonsense questions by manually structuring event sequences logically, followed by programmatic question generation. Concurrently, recognizing the significance of tasks rooted in real-world events, we introduce \textit{facts} problems. These focus on major historical events, spanning from ancient to contemporary times, and are sourced from Wikipedia timelines. Models are posed with challenges such as sequencing: “Arrange the following events in chronological order” and verification queries like, “Is the following sequence of events in the correct chronological order?”. 

\noindent \textbf{Frequency} The frequency task assesses a model's ability to understand how often events take place over time and comprises six distinct categories of problems. For the \textit{commonsense} category, we source questions from the MCTACO dataset related to frequency. Each selected category ensures the presence of at least two incorrect options and one correct one. To prevent models from memorizing specific answer orders, we randomize the placement of the correct answers. In the \textit{reading comprehension} category, questions are chosen from the SQuAD dataset~\citep{rajpurkar2016squad} based on frequency-oriented keywords like “how often”, “how many times”, and “how frequent”. The \textit{application} and \textit{computation} categories are mainly made up of human-curated templates that test the model's ability to infer time intervals and compute either previous or subsequent occurrences. The \textit{comparison} problems blend real and artificially conceived events, challenging the model's ability to differentiate frequency nuances. Lastly, the \textit{facts} category draws questions from various sources, with Wikipedia being the primary one, centering on queries related to events that are known to happen regularly or periodically in either historical or contemporary settings. 

\noindent \textbf{Duration} The duration task is designed to assess a model's capability to comprehend the length of events or periods of time and encompasses seven distinct categories of problems. The \textit{commonsense} problems are derived from the MCTACO dataset, probing the model's fundamental understanding of event durations grounded in everyday scenarios. The extraction methods mirror those used for the “frequency” task. The \textit{reading comprehension} category sources questions from the SQuAD dataset, selecting those with duration-oriented keywords like “how long”, “how many years”, and “how much time”. Apart from the aforementioned subtasks, all other categories consist of human-curated templates or problems. The \textit{analogy inference} category assesses the model's ability to discern durations through analogical reasoning. The \textit{computation} category tests mathematical precision, where problems often require arithmetic operations to determine event durations. Comparative analysis is examined in two subtasks: \textit{direct comparison}, which demands straightforward judgments of event durations involving both real and artificial events; and \textit{multi-step comparison}, which challenges the model to infer and integrate information across sequential statements. Lastly, the \textit{facts} category primarily draws from Wikipedia, furnishing questions anchored in well-documented historical or contemporary durations.

\noindent \textbf{Typical Time} The typical time task is constructed to evaluate a model's understanding of when events or activities typically occur, segmented into four distinct categories. The \textit{commonsense} category draws problems from the MCTACO dataset, exploring the model's innate comprehension of event timings as they manifest in daily scenarios. The extraction method for this subtask is similar to that used for the “frequency” task. The \textit{comparison} category, comprising human-curated statements and problems, delves into relative timing. This category involves determining which of two presented scenarios is more temporally typical or discerning which event customarily precedes the other. The \textit{facts} category, primarily sourced from Wikipedia timelines spanning ancient history to the 21st century, provides the model with specific historical or established events and expects it to identify the precise times or periods associated with them. Lastly, the \textit{reading comprehension} problem sets source questions from the SQuAD dataset. This category filters problems based on keywords like “at what time”, “when did”, and “in what year”, challenging the model to extract specific temporal data from passages.

\subsection{Temporal Interpretation and Computation Tasks}
This group of tasks is fundamental in gauging a model's adeptness at deciphering, processing, and computing temporal information.

\noindent \textbf{Ambiguity Resolution} The temporal ambiguity resolution task aims to gauge a model's ability to decipher and resolve uncertainties related to temporal expressions, divided into five subtasks. The \textit{interpretation} category evaluates the model's comprehension of ambiguous time-related phrases commonly used in everyday language. The \textit{calendar shift} subtask necessitates the conversion between different calendar systems, such as the Julian and Gregorian. The \textit{long-term shift}, \textit{mid-term shift}, and \textit{short-term shift} categories challenge the model's capacity to adjust dates over long (i.e., years), medium (i.e., months, weeks, days), and short (i.e., hours, minutes, seconds) timeframes, respectively. All questions across these categories originate from carefully crafted human templates.

\noindent \textbf{Arithmetic} The temporal arithmetic task evaluates a model's capacity to manage calculations related to time, organized into nine distinct subtasks. The \textit{application} category presents real-world scenarios such as time calculations involving schooling, vacations, homework, and flights. \textit{Date computation} sets focus on adding or subtracting days from specified dates to determine a new date. \textit{hour adjustment} subtasks, divided into \textit{12-hour} and \textit{24-hour} formats, require the model to calculate time differences or additions. The \textit{month shift} subtask examines the model's ability to pinpoint a month that is a certain number of months away from a specified month. The \textit{week identification} problems determine the exact week number within a year based on a given date. In \textit{year shift}, the model discerns a year a certain number of years relative to a provided year. \textit{time computation} evaluates the model's proficiency in converting various time units, especially over shorter durations like days, hours, minutes, and seconds. Lastly, the \textit{time zone conversion} category requires the model to convert times between different zones. Both the question templates and the programs used to formulate answers derive from human expertise.

\subsection{Advanced Temporal and Conceptual Understanding Tasks}
This group of tasks is fundamental in assessing a model's depth of comprehension in time-oriented narratives and in discerning complex conceptual relationships. 

\noindent \textbf{Relation}
The temporal relation task seeks to assess a model's ability to identify the relationship between two entities involving time, categorized as either an \textit{event-to-time} or an \textit{event-to-event} association. Questions are crafted based on the TempEval-3 Silver dataset~\citep{uzzaman2013semeval}. The context sentences, which contain the two entities in question, are directly extracted from the original passages. One inherent challenge of this task lies in the subtle nuances among the fixed set of relations. For instance, distinguishing between relations like “BEFORE” and “IMMEDIATELY BEFORE” can be particularly demanding, as they require fine-grained comprehension of temporal sequences. With the predetermined relations from the dataset, the correct relation option is randomized in its placement, while distractor options are chosen from the pool of remaining relations.

\noindent \textbf{Temporal NLI}
The temporal NLI task is designed to evaluate a model's ability in \textit{natural language inference}, with a particular emphasis on statements that involve temporal elements. We source questions from prevalent NLI datasets, including Stanford Natural Language Inference datasets (SNLI)~\citep{bowman2015large} and Multi-Genre Natural Language Inference (MNLI)~\citep{williams2018broad}. Data from the MNLI dataset includes training and validation sets, while data from the SNLI dataset includes training, validation, and test sets. We select problems based on keywords that capture a range of temporal nuances, such as explicit references (e.g., ‘tomorrow', ‘later'), months (e.g., ‘May', ‘October'), seasons (e.g., ‘summer', ‘winter'), and temporal actions (e.g., ‘in advance', ‘postpone'). Consistent with the original task, the three response options for all questions are: “Entailment”, “Neutral”, and “Contradiction”.

\noindent \textbf{Causality} The temporal causality task assesses a model's capability to discern cause-and-effect relationships within scenarios influenced by time. Drawing inspiration from the Choice of Plausible Alternatives (COPA) dataset~\citep{roemmele2011choice}, we select questions that naturally contain temporal elements such as ‘postpone', ‘tomorrow', ‘summer', and ‘clock'. Additionally, we manually craft problems to highlight the temporal nature of COPA-style questions. Each problem presents a situation that revolves around time, followed by a question pinpointing either the most plausible \textit{cause} or \textit{effect} of that situation. Both options for these problems are carefully created by hand. For augmentation purposes, we create additional, mirrored instances for each original sample. This approach ensures that for a given question with two options, each option is supported by a uniquely tailored premise, effectively creating a distinct and relevant context for both choices.

\noindent \textbf{Storytelling} The \textit{temporal storytelling} task is designed to assess a model's ability to predict the appropriate ending of stories that emphasize temporal elements. We source questions from the ROCStories (ROC)~\citep{mostafazadeh2016corpus} and Story Cloze Test (SCT)~\citep{mostafazadeh2017lsdsem} datasets. We identify and select stories that contain notable temporal components by filtering them using keywords such as ‘now', ‘tomorrow', ‘future', ‘always', and ‘postpone', among others. The typical format of the task presents a story comprising four sentences, followed by two potential endings. The model is required to choose the most appropriate conclusion for the story. In the case of SCT, which inherently provides two endings for each story, our focus remains on selecting stories with evident temporal aspects. To further enrich our dataset, we take the initial four sentences from the ROC and employ GPT-2~\citep{radford2019language} to produce an alternate, incorrect ending, initiated with the prompt “unexpectedly”. Subsequently, we filter this augmented data to ensure that stories emphasize the desired temporal themes.

\section{Experiments}
In our evaluation, we compare the performance of prevalent LLMs across all datasets and analyze the mistakes they make. We report the best results after multiple runs for each experimental setting. 

\subsection{Experimental Setup}
We evaluate the performance of several well-known language models on the TRAM benchmark, which is organized into two main categories. In the first category, we employ four popular LLMs: Llama-2-70b-chat~\citep{touvron2023llama}, Gemini Pro~\citep{anil2023palm}, GPT-3.5 Turbo, and GPT-4 Turbo~\citep{openai2023gpt4}. Each of these models is accessed using its corresponding API key. Specifically, we query Gemini through Google Vertex
AI, the GPT models through the OpenAI API, and
Llama2 via DeepInfra. Following \citep{tan-etal-2023-towards} and considering API cost constraints, we evaluate model performance on 300 randomly selected examples per category from the test set, using all available examples for categories with fewer than 300. For all evaluations, greedy decoding (i.e., temperature = 0) is applied during model response generation. We evaluate each model using two prompting strategies: standard prompting (SP)~\citep{brown2020language, kojima2022large} and CoT~\citep{wei2022chain} prompting. Under both strategies, the models undergo tests in zero-shot and 5-shot settings. In the 5-shot scenario, exemplars are consistently drawn from the development set. Step-by-step answers associated with CoT prompting are obtained through human annotation. More details about prompts can be found in Appendix~\ref{prompts}.

\begin{table*}[h]
\centering 
\caption{Performance comparison of each model across ten tasks in TRAM. GPT-4 consistently outperforms other models under both zero-shot (0S) and 5-shot (5S) settings across the majority of tasks. Interestingly, the RoBERTa-large model achieves a higher average performance than models with larger architectures, such as Llama2. Human expert performance serves as an upper bound, illustrating that there still exists room for improvement in LLMs on TeR tasks. The abbreviations \textit{Freq., Dur., Arith., Rel., Caus.} refer to frequency, duration, arithmetic, relation, and causality, respectively. All values are percentages. Best model results are highlighted in bold.}
\resizebox{\linewidth}{!}{%
\vspace{1cm}
\begin{NiceTabular}{ccccccccccc|c}
    \toprule
    \multirow{2}{*}{\textbf{Model}} & \textbf{Order} & \textbf{Freq.} & \textbf{Dur.} & \textbf{Typical Time } & \textbf{Amb. Res.} & \textbf{Arith.} & \textbf{Rel.} & \textbf{NLI} & \textbf{Caus.} & \textbf{Story} & \textbf{Average}\\ 
    & \textit{Acc.} & \textit{Acc.} & \textit{Acc.} & \textit{Acc.} & \textit{Acc.} & \textit{Acc.} & \textit{Acc./F1} & \textit{Acc./F1} & \textit{Acc.} & \textit{Acc.} \\ 
    \midrule
    Random & 33.3 & 33.3 & 33.3 & 33.3 & 33.3 & 25.0 & 33.3/33.3 & 33.3/33.3 & 50.0 & 50.0 & 35.4\\
    \midrule
    Llama2 (0S, SP) & 51.3 & 73.5 & 64.9 & 74.1 & 46.9 & 52.6 & 35.2/33.1 & 64.4/63.9 & 90.5 & 86.7 & 61.4 \\
    Llama2 (0S, CoT) & 52.9 & 75.4 & 66.3 & 75.5 & 49.4 & 55.6 & 40.1/38.5 & 67.7/67.4 & 92.0 & 88.2 & 64.1 \\
    Llama2 (5S, SP) & 52.2 & 74.1 & 65.7 & 74.6 & 48.0 & 53.9 & 38.1/36.6 & 65.2/64.7 & 92.0 & 87.3 & 62.7\\
    Llama2 (5S, CoT) & 53.8 & 76.3 & 67.1 & 75.9 & 50.7 & 57.8 & 43.0/41.3 & 69.8/69.2 & 93.6 & 88.5 & 65.6 \\
    \midrule
    Gemini (0S, SP) & 55.4 & 86.2 & 83.9 & 82.7 & 75.1 & 69.8 & 60.5/60.1 & 69.5/70.7 & 92.5 & 91.2 & 74.8 \\
    Gemini (0S, CoT) & 56.9 & 87.6 & 84.2 & 83.6 & 76.9 & 71.8 & 64.2/63.6 & 70.9/71.8 & 94.0 & 92.0 & 76.5\\
    Gemini (5S, SP) & 56.4 & 86.5 & 84.5 & 82.9 & 75.8 & 70.4 & 62.8/62.3 & 70.4/71.0 & 94.2 & 91.5 & 75.7 \\
    Gemini (5S, CoT) & 57.4 & 88.2 & 86.3 & 83.8 & 77.4 & 72.5 & 65.1/64.9 & 72.3/73.1 & \textbf{95.3} & 92.2 & 77.4\\
    \midrule
    GPT-3.5 (0S, SP) & 52.5 & 76.3 & 70.8 & 77.8 & 71.6 & 72.8 & 40.5/39.1 & 73.8/74.2 & 93.4 & 90.5 & 69.4 \\
    GPT-3.5 (0S, CoT) & 53.7 & 78.3 & 72.3 & 78.7 & 74.6 & 74.8 & 44.1/42.9 & 75.2/75.7 & 94.5 & 91.7 & 71.4\\
    GPT-3.5 (5S, SP) & 53.2 & 77.8 & 71.6 & 79.2 & 73.4 & 73.7 & 42.5/41.3 & 74.5/75.0 & 94.5 & 91.0 & 70.6 \\
    GPT-3.5 (5S, CoT) & 54.8 & 79.2 & 72.7 & 80.3 & 75.2 & 75.0 & 45.9/45.2 & 76.3/76.9 & 94.8 & 91.7 & 72.3 \\
    \midrule
    GPT-4 (0S, SP) & 64.7 & 85.2 & 86.1 & 84.6 & 82.3 & 87.1 & 60.6/58.8 & 82.9/85.3 & 92.6 & 91.0 & 80.1\\
    GPT-4 (0S, CoT) & 66.2 & 87.7 & 86.4 & 85.5 & 84.1 & 88.9 & 63.6/62.9 & 85.4/87.2 & 92.9 & 93.2 & 82.0\\
    GPT-4 (5S, SP) & 65.8 & 86.3 & 87.3 & 84.8 & 83.6 & 88.3 & 62.0/61.5 & 83.7/86.4 & 92.6 & 91.6 & 81.2 \\
    GPT-4 (5S, CoT) & \textbf{69.5} & \textbf{90.7} & \textbf{89.2} & \textbf{87.2} & \textbf{87.1} & \textbf{91.2} & 66.5/65.2 & \textbf{87.7/89.6} & 95.0 & \textbf{93.6} & \textbf{84.4} \\
    \midrule
    BERT-base & 50.0 & 47.3 & 50.0 & 53.0 & 36.6 & 25.9 & 86.5/86.6 & 53.0/53.4 & 81.0 & 79.0 & 58.5\\
    BERT-large & 52.5 & 53.1 & 53.3 & 56.8 & 37.4 & 28.3 & 89.5/89.5 & 59.5/60.1 & 85.0 & 81.3 & 62.2\\
    RoBERTa-base & 50.8 & 54.5 & 51.8 & 55.3 & 37.4 & 26.4 & 87.0/86.8 & 64.5/64.9 & 82.3 & 81.3 & 61.9 \\
    RoBERTa-large & 55.5 & 57.7 & 55.4 & 60.0 & 41.0 & 29.1 & 90.0/90.0 & 70.0/70.3 & 88.0 & 84.0 & 65.9\\
    RST & 54.5 & 56.2 & 52.3 & 58.7	& 39.8 & 31.6 & \textbf{91.5/91.6} &	68.2/68.7 & 87.5 & 82.2 & 65.2 \\
    \midrule
    Human Experts & 86.0 & 96.3 & 97.7 & 94.5 & 94.8 & 98.7 & 96.0/96.0 & 92.0/92.4 & 100.0 & 98.0 & 95.2\\
  
    \bottomrule
\end{NiceTabular}}
\label{tab: overall_performance}
\end{table*}

In the second category, we consider minimal supervision as opposed to traditional fully supervised learning to establish baseline evaluations. The rationale behind this decision is driven by the intention to leverage the inherent world knowledge of the models and to ensure an equitable comparison with the previously mentioned LLMs. We employ four representative BERT-style models, including BERT-base, BERT-large~\citep{kenton2019bert}, RoBERTa-base, and RoBERTa-large~\citep{liu2019roberta}. For temporal NLI, we employ the Sequence Classification versions of BERT and RoBERTa from Huggingface, which align with the task demands. For other tasks, we use their Multiple Choice configurations. Additionally, we include the RST~\citep{yuan2022restructured}, a domain-specific TeR model, to benchmark against the generalist models. The data sampling strategy for minimal supervision is structured based on the size of the original dataset. For datasets with around 1k samples, we randomly
select 50\% of the remaining data after setting aside the test data used for LLM evaluation. For
datasets with sizes between 3k and 10k, we select 10\%. For those with sizes between 10k and 100k, we sample 2.5\%, and for datasets with more than 100k examples, we take 1\%. This limited training data is used for model fine-tuning. The same test set is used consistently with LLMs.

In addition to evaluating model performance, multiple expert annotators worked on each problem type for every task in TRAM to better understand human performance. Each expert answered a subset of the 50 questions from each category of every task, which were randomly selected from the test set. Collectively, they tackled about 1,900 questions across TRAM. Further details on human expert annotators and human non-specialists are provided in Appendix~\ref{human_expertise}.

\begin{figure*}[ht]
\centering
\includegraphics[width=\textwidth]{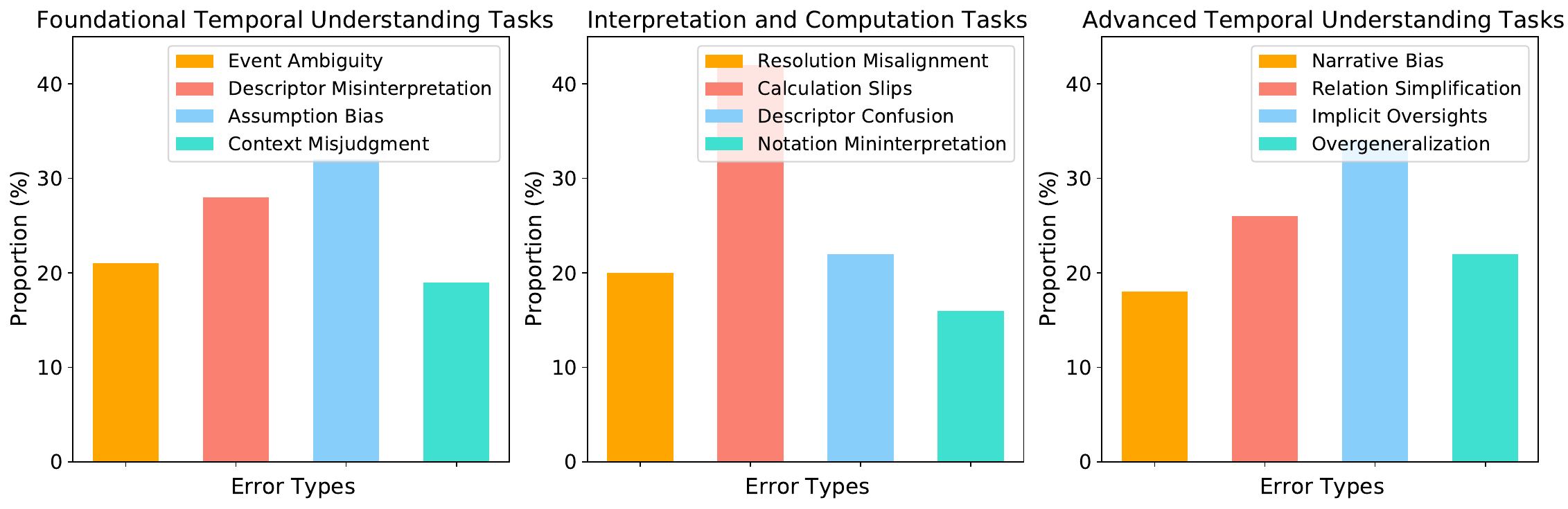}  
\caption{Error type distribution for three groups of tasks in TRAM. Models often struggle with subtle details and hidden clues across all categories.}
\label{fig: errors}
\end{figure*}

\subsection{Overall Performance Comparison}
We compare the performance of different models across ten tasks, as shown in Table~\ref{tab: overall_performance}. There are several key takeaways.
First, GPT-4 consistently outperforms other models across the majority of tasks, demonstrating a performance advantage of over 9\% compared to the second-best model, Gemini, under 5-shot CoT. Second, CoT often results in performance enhancements, corroborating the findings from~\citep{wei2022chain} and emphasizing the efficacy of step-by-step prompting in augmenting LLMs' performance in intricate reasoning tasks. Third, it is notable that RoBERTa-large, despite its size, surpasses the larger Llama2 in average performance. This observation underscores that sheer model size does not always equate to superior performance. Several factors might contribute to this outcome. RoBERTa-large may utilize optimization strategies particularly beneficial for these tasks. Additionally, inherent features or efficiencies in its architecture might enhance its ability to understand and process temporal cues. Delving deeper into task-specific performance, certain tasks such as ambiguity resolution and arithmetic show considerable variance across models. For LLMs, performance on the arithmetic task varies significantly, ranging from 52.6\% to 91.2\%. Moreover, BERT and RoBERTa exhibit exceptional performance in the temporal relation task, potentially due to their bidirectional contextual processing and emphasis on token-level relationships. This contrasts sharply with their average or below-average performance in other tasks. For the specialized RST model, we observe comparable average performance with RoBERTa-large, indicating the benefits of tailored training for domain-specific tasks. The discrepancies in performance among models suggest that certain architectures or training methodologies are better suited for specific types of reasoning or tasks, highlighting the need for tailored approaches. Finally, despite the lead of GPT-4, it remains 12.9\% behind human performance, underscoring the potential for further LLM enhancements.

\subsection{Error Analysis}
To better understand the mistakes made by models, we manually analyze instances where a model has made incorrect choices or provided inappropriate answers, focusing exclusively on LLMs. Figure~\ref{fig: errors} illustrates the common error types and their proportions for each task group. In foundational temporal understanding tasks, “assumption bias” was the most frequent error, accounting for 32\% of all mistakes. In the interpretation and computation tasks, “calculation slips” dominated, making up 42\% of the errors. “Implicit oversights” led in the advanced temporal understanding tasks with a representation of 34\%. Detailed descriptions of each error type can be found in Appendix~\ref{error_types}.

\section{Discussion}
We introduce TRAM, a comprehensive benchmark spanning ten diverse tasks, to evaluate the temporal reasoning of LLMs. The contrasting performances across models emphasize the significance of experimental strategies and shed light on the intrinsic challenges. This benchmark serves as a tool for researchers to identify model limitations and guide further advancements in this domain.

\noindent \textbf{Implications of TRAM} The introduction of TRAM establishes a new paradigm for probing the temporal reasoning capabilities of LLMs. Unlike previous benchmarks, which often offered fragmented insights into temporal tasks, TRAM provides a comprehensive system. This allows for a unified evaluation of how models comprehend both rudimentary temporal concepts and complex temporal narratives. The differentiation in task complexity within TRAM elucidates the various stages of temporal understanding. In particular, TRAM underscores challenges like decoding implicit temporal cues and navigating intricate temporal relationships, providing a roadmap for future improvements in LLMs in this area.

\noindent \textbf{Future Directions} TRAM has initiated a step towards evaluating LLMs' temporal reasoning capabilities, but there are further avenues to explore. Going forward, we will experiment with more test data and refine tailored prompting techniques for each task through iterative testing. Moreover, we plan to expand the benchmark to include varied question formats. For generative tasks, this might encompass short answers and summarization. Even within MCQs, we intend to incorporate questions that may have one or more correct answers, allowing for a more comprehensive evaluation. We also plan to fine-tune leading open-source LLMs, such as Llama3, Phi-3, and Gemma, on these tasks to develop domain-specific expert models. These efforts aim to create tailored LLMs that can more effectively understand and reason about time across various contexts.

\section{Limitations}
While TRAM sets a holistic standard for TeR assessment, we acknowledge its limitations. One primary concern is the subset evaluation of the test set, which may not reflect the full spectrum of LLMs' TeR capabilities. Furthermore, the MCQ format may allow LLMs to guess randomly, skewing performance evaluations. Moreover, textual questions may not capture the entire complexity of TeR tasks, as real-world scenarios often integrate multi-modal cues such as images and videos.

\bibliography{custom}

\begin{thebibliography}{34}
\expandafter\ifx\csname natexlab\endcsname\relax\def\natexlab#1{#1}\fi

\bibitem[{Anil et~al.(2023)Anil, Dai, Firat, Johnson, Lepikhin, Passos, Shakeri, Taropa, Bailey, Chen et~al.}]{anil2023palm}
Rohan Anil, Andrew~M Dai, Orhan Firat, Melvin Johnson, Dmitry Lepikhin, Alexandre Passos, Siamak Shakeri, Emanuel Taropa, Paige Bailey, Zhifeng Chen, et~al. 2023.
\newblock Palm 2 technical report.
\newblock \emph{arXiv preprint arXiv:2305.10403}.

\bibitem[{Bowman et~al.(2015)Bowman, Angeli, Potts, and Manning}]{bowman2015large}
Samuel~R Bowman, Gabor Angeli, Christopher Potts, and Christopher~D Manning. 2015.
\newblock A large annotated corpus for learning natural language inference.
\newblock In \emph{Proceedings of the 2015 Conference on Empirical Methods in Natural Language Processing}. Association for Computational Linguistics.

\bibitem[{Brown et~al.(2020)Brown, Mann, Ryder, Subbiah, Kaplan, Dhariwal, Neelakantan, Shyam, Sastry, Askell et~al.}]{brown2020language}
Tom Brown, Benjamin Mann, Nick Ryder, Melanie Subbiah, Jared~D Kaplan, Prafulla Dhariwal, Arvind Neelakantan, Pranav Shyam, Girish Sastry, Amanda Askell, et~al. 2020.
\newblock Language models are few-shot learners.
\newblock \emph{Advances in neural information processing systems}, 33:1877--1901.

\bibitem[{Chen et~al.(2021)Chen, Wang, and Wang}]{chen2021dataset}
Wenhu Chen, Xinyi Wang, and William~Yang Wang. 2021.
\newblock A dataset for answering time-sensitive questions.
\newblock \emph{arXiv preprint arXiv:2108.06314}.

\bibitem[{Dhingra et~al.(2022)Dhingra, Cole, Eisenschlos, Gillick, Eisenstein, and Cohen}]{dhingra2022time}
Bhuwan Dhingra, Jeremy~R Cole, Julian~Martin Eisenschlos, Daniel Gillick, Jacob Eisenstein, and William~W Cohen. 2022.
\newblock Time-aware language models as temporal knowledge bases.
\newblock \emph{Transactions of the Association for Computational Linguistics}, 10:257--273.

\bibitem[{Hendrycks et~al.(2020)Hendrycks, Burns, Basart, Zou, Mazeika, Song, and Steinhardt}]{hendrycks2020measuring}
Dan Hendrycks, Collin Burns, Steven Basart, Andy Zou, Mantas Mazeika, Dawn Song, and Jacob Steinhardt. 2020.
\newblock Measuring massive multitask language understanding.
\newblock In \emph{International Conference on Learning Representations}.

\bibitem[{Kenton and Toutanova(2019)}]{kenton2019bert}
Jacob Devlin Ming-Wei~Chang Kenton and Lee~Kristina Toutanova. 2019.
\newblock Bert: Pre-training of deep bidirectional transformers for language understanding.
\newblock In \emph{Proceedings of naacL-HLT}, volume~1, page~2.

\bibitem[{Kojima et~al.(2022)Kojima, Gu, Reid, Matsuo, and Iwasawa}]{kojima2022large}
Takeshi Kojima, Shixiang~Shane Gu, Machel Reid, Yutaka Matsuo, and Yusuke Iwasawa. 2022.
\newblock Large language models are zero-shot reasoners.
\newblock \emph{Advances in neural information processing systems}, 35:22199--22213.

\bibitem[{Liu et~al.(2019)Liu, Ott, Goyal, Du, Joshi, Chen, Levy, Lewis, Zettlemoyer, and Stoyanov}]{liu2019roberta}
Yinhan Liu, Myle Ott, Naman Goyal, Jingfei Du, Mandar Joshi, Danqi Chen, Omer Levy, Mike Lewis, Luke Zettlemoyer, and Veselin Stoyanov. 2019.
\newblock Roberta: A robustly optimized bert pretraining approach.
\newblock \emph{arXiv preprint arXiv:1907.11692}.

\bibitem[{Min et~al.(2021)Min, Ross, Sulem, Veyseh, Nguyen, Sainz, Agirre, Heintz, and Roth}]{min2021recent}
Bonan Min, Hayley Ross, Elior Sulem, Amir Pouran~Ben Veyseh, Thien~Huu Nguyen, Oscar Sainz, Eneko Agirre, Ilana Heintz, and Dan Roth. 2021.
\newblock Recent advances in natural language processing via large pre-trained language models: A survey.
\newblock \emph{ACM Computing Surveys}.

\bibitem[{Mostafazadeh et~al.(2016)Mostafazadeh, Chambers, He, Parikh, Batra, Vanderwende, Kohli, and Allen}]{mostafazadeh2016corpus}
Nasrin Mostafazadeh, Nathanael Chambers, Xiaodong He, Devi Parikh, Dhruv Batra, Lucy Vanderwende, Pushmeet Kohli, and James Allen. 2016.
\newblock A corpus and cloze evaluation for deeper understanding of commonsense stories.
\newblock In \emph{Proceedings of the 2016 Conference of the North American Chapter of the Association for Computational Linguistics: Human Language Technologies}, pages 839--849.

\bibitem[{Mostafazadeh et~al.(2017)Mostafazadeh, Roth, Louis, Chambers, and Allen}]{mostafazadeh2017lsdsem}
Nasrin Mostafazadeh, Michael Roth, Annie Louis, Nathanael Chambers, and James Allen. 2017.
\newblock Lsdsem 2017 shared task: The story cloze test.
\newblock In \emph{Proceedings of the 2nd Workshop on Linking Models of Lexical, Sentential and Discourse-level Semantics}, pages 46--51.

\bibitem[{OpenAI(2023)}]{openai2023gpt4}
OpenAI. 2023.
\newblock \href {http://arxiv.org/abs/2303.08774} {Gpt-4 technical report}.

\bibitem[{Pustejovsky et~al.(2003)Pustejovsky, Hanks, Sauri, See, Gaizauskas, Setzer, Radev, Sundheim, Day, Ferro et~al.}]{pustejovsky2003timebank}
James Pustejovsky, Patrick Hanks, Roser Sauri, Andrew See, Robert Gaizauskas, Andrea Setzer, Dragomir Radev, Beth Sundheim, David Day, Lisa Ferro, et~al. 2003.
\newblock The timebank corpus.
\newblock In \emph{Corpus linguistics}, volume 2003, page~40. Lancaster, UK.

\bibitem[{Radford et~al.(2019)Radford, Wu, Child, Luan, Amodei, Sutskever et~al.}]{radford2019language}
Alec Radford, Jeffrey Wu, Rewon Child, David Luan, Dario Amodei, Ilya Sutskever, et~al. 2019.
\newblock Language models are unsupervised multitask learners.

\bibitem[{Rajpurkar et~al.(2016)Rajpurkar, Zhang, Lopyrev, and Liang}]{rajpurkar2016squad}
Pranav Rajpurkar, Jian Zhang, Konstantin Lopyrev, and Percy Liang. 2016.
\newblock Squad: 100,000+ questions for machine comprehension of text.
\newblock In \emph{Proceedings of the 2016 Conference on Empirical Methods in Natural Language Processing}, pages 2383--2392.

\bibitem[{Roemmele et~al.(2011)Roemmele, Bejan, and Gordon}]{roemmele2011choice}
Melissa Roemmele, Cosmin~Adrian Bejan, and Andrew~S Gordon. 2011.
\newblock Choice of plausible alternatives: An evaluation of commonsense causal reasoning.
\newblock In \emph{2011 AAAI Spring Symposium Series}.

\bibitem[{Tan et~al.(2023)Tan, Ng, and Bing}]{tan-etal-2023-towards}
Qingyu Tan, Hwee~Tou Ng, and Lidong Bing. 2023.
\newblock Towards benchmarking and improving the temporal reasoning capability of large language models.
\newblock In \emph{Proceedings of the 61st Annual Meeting of the Association for Computational Linguistics (Volume 1: Long Papers)}, pages 14820--14835.

\bibitem[{Team et~al.(2023)Team, Anil, Borgeaud, Wu, Alayrac, Yu, Soricut, Schalkwyk, Dai, Hauth et~al.}]{team2023gemini}
Gemini Team, Rohan Anil, Sebastian Borgeaud, Yonghui Wu, Jean-Baptiste Alayrac, Jiahui Yu, Radu Soricut, Johan Schalkwyk, Andrew~M Dai, Anja Hauth, et~al. 2023.
\newblock Gemini: a family of highly capable multimodal models.
\newblock \emph{arXiv preprint arXiv:2312.11805}.

\bibitem[{Touvron et~al.(2023)Touvron, Martin, Stone, Albert, Almahairi, Babaei, Bashlykov, Batra, Bhargava, Bhosale et~al.}]{touvron2023llama}
Hugo Touvron, Louis Martin, Kevin Stone, Peter Albert, Amjad Almahairi, Yasmine Babaei, Nikolay Bashlykov, Soumya Batra, Prajjwal Bhargava, Shruti Bhosale, et~al. 2023.
\newblock Llama 2: Open foundation and fine-tuned chat models.
\newblock \emph{arXiv preprint arXiv:2307.09288}.

\bibitem[{UzZaman et~al.(2013)UzZaman, Llorens, Derczynski, Allen, Verhagen, and Pustejovsky}]{uzzaman2013semeval}
Naushad UzZaman, Hector Llorens, Leon Derczynski, James Allen, Marc Verhagen, and James Pustejovsky. 2013.
\newblock Semeval-2013 task 1: Tempeval-3: Evaluating time expressions, events, and temporal relations.
\newblock In \emph{Second Joint Conference on Lexical and Computational Semantics (* SEM), Volume 2: Proceedings of the Seventh International Workshop on Semantic Evaluation (SemEval 2013)}, pages 1--9.

\bibitem[{Wang et~al.(2022{\natexlab{a}})Wang, Wei, Schuurmans, Le, Chi, Narang, Chowdhery, and Zhou}]{wang2022self}
Xuezhi Wang, Jason Wei, Dale Schuurmans, Quoc Le, Ed~Chi, Sharan Narang, Aakanksha Chowdhery, and Denny Zhou. 2022{\natexlab{a}}.
\newblock Self-consistency improves chain of thought reasoning in language models.
\newblock \emph{arXiv preprint arXiv:2203.11171}.

\bibitem[{Wang and Zhao(2023)}]{wang2023metacognitive}
Yuqing Wang and Yun Zhao. 2023.
\newblock Metacognitive prompting improves understanding in large language models.
\newblock \emph{arXiv preprint arXiv:2308.05342}.

\bibitem[{Wang et~al.(2022{\natexlab{b}})Wang, Zhao, Callcut, and Petzold}]{wang2022integrating}
Yuqing Wang, Yun Zhao, Rachael Callcut, and Linda Petzold. 2022{\natexlab{b}}.
\newblock Integrating physiological time series and clinical notes with transformer for early prediction of sepsis.
\newblock \emph{arXiv preprint arXiv:2203.14469}.

\bibitem[{Wang et~al.(2022{\natexlab{c}})Wang, Zhao, and Petzold}]{wang2022enhancing}
Yuqing Wang, Yun Zhao, and Linda Petzold. 2022{\natexlab{c}}.
\newblock Enhancing transformer efficiency for multivariate time series classification.
\newblock \emph{arXiv preprint arXiv:2203.14472}.

\bibitem[{Wang et~al.(2023)Wang, Zhao, and Petzold}]{wang2023large}
Yuqing Wang, Yun Zhao, and Linda Petzold. 2023.
\newblock Are large language models ready for healthcare? a comparative study on clinical language understanding.
\newblock \emph{arXiv preprint arXiv:2304.05368}.

\bibitem[{Wei et~al.(2022)Wei, Wang, Schuurmans, Bosma, Xia, Chi, Le, Zhou et~al.}]{wei2022chain}
Jason Wei, Xuezhi Wang, Dale Schuurmans, Maarten Bosma, Fei Xia, Ed~Chi, Quoc~V Le, Denny Zhou, et~al. 2022.
\newblock Chain-of-thought prompting elicits reasoning in large language models.
\newblock \emph{Advances in Neural Information Processing Systems}, 35:24824--24837.

\bibitem[{Wei et~al.(2023)Wei, Su, Ma, Yu, Lei, Zhang, Zhao, and Liu}]{wei2023menatqa}
Yifan Wei, Yisong Su, Huanhuan Ma, Xiaoyan Yu, Fangyu Lei, Yuanzhe Zhang, Jun Zhao, and Kang Liu. 2023.
\newblock Menatqa: A new dataset for testing the temporal comprehension and reasoning abilities of large language models.
\newblock In \emph{Findings of the Association for Computational Linguistics: EMNLP 2023}, pages 1434--1447.

\bibitem[{Williams et~al.(2018)Williams, Nangia, and Bowman}]{williams2018broad}
Adina Williams, Nikita Nangia, and Samuel Bowman. 2018.
\newblock A broad-coverage challenge corpus for sentence understanding through inference.
\newblock In \emph{Proceedings of the 2018 Conference of the North American Chapter of the Association for Computational Linguistics: Human Language Technologies, Volume 1 (Long Papers)}, pages 1112--1122.

\bibitem[{Yao et~al.(2023)Yao, Yu, Zhao, Shafran, Griffiths, Cao, and Narasimhan}]{yao2023tree}
Shunyu Yao, Dian Yu, Jeffrey Zhao, Izhak Shafran, Thomas~L Griffiths, Yuan Cao, and Karthik Narasimhan. 2023.
\newblock Tree of thoughts: Deliberate problem solving with large language models.
\newblock \emph{arXiv preprint arXiv:2305.10601}.

\bibitem[{Yuan and Liu(2022)}]{yuan2022restructured}
Weizhe Yuan and Pengfei Liu. 2022.
\newblock restructured pre-training.
\newblock \emph{arXiv preprint arXiv:2206.11147}.

\bibitem[{Zhao et~al.(2023)Zhao, Zhou, Li, Tang, Wang, Hou, Min, Zhang, Zhang, Dong et~al.}]{zhao2023survey}
Wayne~Xin Zhao, Kun Zhou, Junyi Li, Tianyi Tang, Xiaolei Wang, Yupeng Hou, Yingqian Min, Beichen Zhang, Junjie Zhang, Zican Dong, et~al. 2023.
\newblock A survey of large language models.
\newblock \emph{arXiv preprint arXiv:2303.18223}.

\bibitem[{Zhao et~al.(2021)Zhao, Wang, Liu, Xia, Xu, Hong, Zhou, and Petzold}]{zhao2021empirical}
Yun Zhao, Yuqing Wang, Junfeng Liu, Haotian Xia, Zhenni Xu, Qinghang Hong, Zhiyang Zhou, and Linda Petzold. 2021.
\newblock Empirical quantitative analysis of covid-19 forecasting models.
\newblock In \emph{2021 International Conference on Data Mining Workshops (ICDMW)}, pages 517--526. IEEE.

\bibitem[{Zhou et~al.(2019)Zhou, Khashabi, Ning, and Roth}]{zhou2019going}
Ben Zhou, Daniel Khashabi, Qiang Ning, and Dan Roth. 2019.
\newblock “going on a vacation” takes longer than “going for a walk”: A study of temporal commonsense understanding.
\newblock In \emph{Proceedings of the 2019 Conference on Empirical Methods in Natural Language Processing and the 9th International Joint Conference on Natural Language Processing (EMNLP-IJCNLP)}, pages 3363--3369.

\end{thebibliography}

\appendix

\section{Datasets}~\label{data}
This section presents the datasheet for TRAM, a glossary of definitions for all subtasks, and details the dataset construction process, including human-crafted templates for programmatic question generation and the use of temporal keywords to filter questions from existing datasets. Additional example questions for each task, as well as those sourced from existing datasets, are provided. Furthermore, for tasks comprising multiple subtasks, we provide their distribution. Note that the following templates do not represent the full spectrum of templates we used when constructing the datasets.

\subsection{Datasheet for TRAM}

\begin{center}
\fbox{\textbf{OVERVIEW}}
\end{center}

\noindent \textbf{Motivation and Intended Uses.}

\noindent {\color{violet}{1. What are the intended purposes for this benchmark?}}

\noindent The benchmark is designed to establish a standard for evaluating temporal reasoning in large language models. It focuses on three key areas: Foundational Temporal Understanding (such as Duration and Frequency), Temporal Interpretation and Computation (including Ambiguity Resolution and Arithmetic), and Advanced Temporal and Conceptual Understanding (encompassing areas like Causality and Storytelling).

\noindent {\color{violet}{2. Was it designed to address a specific task or fill a particular gap in research or application?}}

\noindent The benchmark is curated to address the need for a robust and comprehensive tool, specifically designed to evaluate temporal reasoning in large language models. It provides a diverse set of tasks that challenge models in the more intricate aspects of temporal reasoning.

\vspace{5mm}

\noindent \textbf{Limitations and Inappropriate Uses.}

\noindent {\color{violet}{3. Are there any specific tasks or applications for which this benchmark should not be used?}}

\noindent The focus of the benchmark is on understanding and interpreting time-related concepts. Therefore, it may not be suitable for evaluations that significantly diverge from temporal reasoning, such as tasks involving texts that require contextual emotional intelligence, or domain-specific applications in medical or legal document analysis.

\begin{center}
\fbox{\textbf{DETAILS}}
\end{center}

\noindent \textbf{Composition.}

\noindent {\color{violet}{4. What do the instances that comprise the benchmark represent?}}

\noindent The instances consist of multiple-choice questions, created from a combination of existing datasets and human-curated problems, with a focus on temporal reasoning tasks. Each instance is specifically designed to assess a language model's ability to process and reason about time in natural language.

\noindent {\color{violet}{5. How many instances are there in total (of each type, if appropriate)?}}

\noindent There are a total of 526,668 problems. Specifically, the dataset comprises 10 main tasks and 38 subtasks. The number of problems for each main task is as follows: Ordering (29,462), Frequency (4,658), Duration (7,232), Typical Time (13,018), Ambiguity Resolution (3,649), Arithmetic (15,629), Relation (102,462), Temporal NLI (282,144), Causality (1,200), and Storytelling (67,214).

\noindent {\color{violet}{6. Does the benchmark contain all possible instances or is it a sample (not necessarily random) of instances from a larger set?}}

\noindent Part of the benchmark comprises a curated selection of instances, representing a comprehensive but not exhaustive collection of temporal reasoning problems. Specifically, it includes problems selectively sourced from existing datasets that exemplify a wide array of temporal reasoning scenarios. Human expertise has verified and determined the representativeness of the selected problems.

\noindent {\color{violet}{7. Is there a label or target associated with each instance?}}

\noindent Yes, the label for each instance is the correct answer to the multiple-choice question, indicated as either A, B, C, or D, and this varies by task.

\noindent {\color{violet}{8. Is the benchmark self-contained, or does it link to or otherwise rely on external resources (e.g., websites, tweets, other datasets)?}}

\noindent The benchmark is partially self-contained. Problems derived from existing datasets have been integrated into TRAM in a way that makes them standalone. This integration includes manually adding distracting or confusing options, filtering out irrelevant questions for relevance, and reformulating problems. For transparency, references are provided for problems that originated from existing data. The remaining questions are heavily driven by human curation, supplemented by programmatic generation.

\noindent {\color{violet}{9. Does the benchmark contain data that might be considered sensitive in any way?}}

\noindent The benchmark does not contain any sensitive data.

\vspace{5mm}
\noindent \textbf{Data Quality.}

\noindent {\color{violet}{10. Is there any missing information in the benchmark?}}

\noindent Everything is included. No data is missing.

\noindent {\color{violet}{11. What errors, sources of noise, or redundancies are important for benchmark users to be aware of?}}

\noindent Firstly, some problems in the benchmark might contain contextual ambiguities leading to multiple plausible interpretations. The benchmark is designed to have one correct answer per question, with the final unique correct answer determined or verified by a group of professionals. Secondly, within the same main task, there may be similar problems with nuanced differences. While complete redundancy of problems across the entire benchmark is avoided, the presence of similar problems is not. Finally, for problems sourced from existing datasets, irrelevant or diverging options may occur during reformulation due to issues with the source data. Further verification checks will be conducted to minimize any errors or noise that may arise in the benchmark.

\noindent {\color{violet}{12. How was the data validated/verified?}}

\noindent The benchmark was initially verified by multiple professionals possessing advanced degrees (M.S. or Ph.D.) in cognitive science and psychology, who provided insights into the nuances of human temporal cognition, as well as in statistics, mathematics, and computer science, for their expertise in analytical rigor required by many tasks. They reviewed the problems for relevance and common errors, such as formatting inconsistencies or logical discrepancies in questions and answers. The final review of the benchmark was conducted by the authors of the TRAM paper, who checked for relevance and removed any obvious noise and redundancies.

\vspace{5mm}
\noindent \textbf{Pre-Processing, Cleaning, and Labeling.}

\noindent {\color{violet}{13. What pre-processing, cleaning, and/or labeling was done on this benchmark?}}

\noindent In the preparation of the benchmark, several key steps were undertaken to ensure its overall quality and relevance:
\begin{enumerate}
\item[1)] Pre-processing: This step involved standardizing the format of problems sourced from relevant existing datasets to align with the TRAM benchmark’s structure. It included unifying the formats of questions and answers, normalizing temporal expressions, and ensuring consistency in language and style. Additionally, over 100k problems in the benchmark were manually crafted, supplemented by program generation.
\item[2)] Cleaning: A thorough review was conducted to identify and correct any obvious errors in the data. This process involved resolving typos, rectifying factual inaccuracies, and eliminating ambiguous or misleading phrasing in both questions and options. However, nuanced errors such as acceptable bias in multiple interpretations of the same problem and subtle logical errors might be overlooked and could still be present in the current version of the benchmark.
\item[3)] Labeling: Each problem in the benchmark was carefully labeled with the correct answer. In the case of multiple-choice questions, plausible distractors were also manually created and added. Labels were verified for accuracy by subject matter experts to ensure that they correctly represented the intended temporal reasoning challenge. 
\end{enumerate}

\noindent {\color{violet}{14. Provide a link to the code used to pre-process/clean/label the data, if available.}}

\noindent The code for data pre-processing is available on the official GitHub page.

\noindent {\color{violet}{15. If there are any recommended data splits (e.g., training, validation, testing), please explain.}}

\noindent For each main task, there is a few-shot development set, with 5 questions per category (subtask), and a separate test set for evaluation.

\vspace{2mm}
\fbox{
  \begin{minipage}{.43\textwidth}
    \centering
    \textbf{ADDITIONAL DETAILS ON} \\
    \textbf{DISTRIBUTION AND MAINTENANCE}
  \end{minipage}
}

\vspace{2mm}
\noindent \textbf{Distribution.}

\noindent {\color{violet}{16. Will the benchmark be distributed to third parties outside of the entity (e.g., company, institution, organization) on behalf of which the dataset was created?}}

\noindent Yes, the benchmark will be publicly available on the Internet.

\noindent {\color{violet}{17. How will the benchmark be distributed (e.g., tarball on website, API, GitHub)?}}

\noindent The benchmark is distributed via the official GitHub page.

\noindent {\color{violet}{18. When will the benchmark be distributed?}}

\noindent The benchmark was first released in September 2023.

\vspace{5mm}
\noindent \textbf{Maintenance.}

\noindent {\color{violet}{19. Who will be supporting/hosting/maintaining the benchmark?}}

\noindent The first author of the TRAM paper will be supporting and maintaining the benchmark. 

\noindent {\color{violet}{20. Will the benchmark be updated (e.g., to correct labeling errors, add new instances, delete instances)?}}

\noindent Updates to question sets, error corrections, and results will be shared on the official GitHub page.

\noindent {\color{violet}{21. Will older versions of the benchmark continue to be supported/hosted/maintained?}}

\noindent Given any updates to the benchmark, older versions will be retained for consistency.

\noindent {\color{violet}{22. If others want to extend/augment/build on/contribute to the benchmark, is there a mechanism for them to do so?}}

\noindent Others wishing to do so should contact the original authors of TRAM about incorporating fixes or extensions.

\subsection{Task Glossary Definitions}
We provide a glossary with definitions of all tasks and subtasks encompassed within our TRAM benchmark for clarity. In our actual dataset formatting, the subcategory (if a task comprises multiple subtasks) or source (if a single subtask is sourced from an existing dataset) is marked for verification and convenient lookup.

\noindent \textbf{Ordering}: Chronological arrangement of events.
\begin{itemize}
\item \textit{Commonsense}: Logical sequencing of events based on general knowledge.
\item \textit{Facts}: Accurate ordering of historical events based on factual information.
\end{itemize}

\noindent \textbf{Frequency}: Determination of how often events occur over time.
\begin{itemize}
\item \textit{Commonsense}: Assessment of event occurrence rates based on general knowledge.
\item \textit{Reading Comprehension}: Frequency information extraction from passages.
\item \textit{Application}: Inference of time intervals and event frequencies.
\item \textit{Computation}: Calculation of event occurrences and intervals.
\item \textit{Comparison}: Differentiation of event frequencies in various contexts.
\item \textit{Facts}: Identification of periodically occurring events.
\end{itemize}

\noindent \textbf{Duration}: Determination of the length of events or time periods.
\begin{itemize}
\item \textit{Commonsense}: Evaluation of time spans in everyday life scenarios.
\item \textit{Reading Comprehension}: Duration information extraction from passages.
\item \textit{Analogy Inference}: Discernment of relative time spans through contextual comparison.
\item \textit{Computation}: Calculation of event lengths.
\item \textit{Direct Comparison}: Straightforward assessment of event durations in a given set.
\item \textit{Multi-step Comparison}: Analysis of relative durations using layered information.
\item \textit{Facts}: Identification of length of factual events.
\end{itemize}

\noindent \textbf{Typical Time}: Determination of when events or activities typically occur.
\begin{itemize}
\item \textit{Commonsense}: Analysis of usual event timings in daily life scenarios.
\item \textit{Comparison}: Assessment of relative event timings and typical sequences.
\item \textit{Facts}: Identification of historical times or periods from established events.
\item \textit{Reading Comprehension}: Specific time information extraction from passages.
\end{itemize}

\noindent \textbf{Ambiguity Resolution}: Resolution of uncertainties in temporal expressions.
\begin{itemize}
\item \textit{Interpretation}: Understanding of ambiguous time-related phrases.
\item \textit{Calendar shift}: Conversion between different calendar systems.
\item \textit{Long-term shift}: Adjustment of dates over extended periods (years).
\item \textit{Mid-term shift}: Date adjustments over intermediate periods (months, weeks, days).
\item \textit{Short-term shift}: Time adjustments over brief periods (hours, minutes, seconds).
\end{itemize}

\noindent \textbf{Arithmetic}: Execution of time-related calculations.
\begin{itemize}
\item \textit{Application}: Real-world time calculation scenarios (schooling, vacations, etc.).
\item \textit{Date Computation}: Addition or subtraction of days to find new dates.
\item \textit{12-hour Adjustment}: Time difference calculations in 12-hour format.
\item \textit{24-hour Adjustment}: Time difference calculations in 24-hour format.
\item \textit{Month Shift}: Identification of a future or past month from a given date.
\item \textit{Week Identification}: Determination of week numbers within a year.
\item \textit{Year Shift}: Calculation of future or past years from a specified year.
\item \textit{Time Computation}: Calculating future or past years from a specified year.
\item \textit{Time Zone Conversion}: Conversion of times between different time zones.
\end{itemize}

\noindent \textbf{Relation}: Identification of the temporal relationship between two entities, either as an event-to-time or event-to-event association.

\noindent \textbf{Temporal NLI}: Assessment of a ‘hypothesis' as true (entailment), false (contradiction), or undetermined (neutral) relative to a ‘premise' with temporal elements.

\noindent \textbf{Causality}: Analysis of cause-and-effect relationships in time-related scenarios.
\begin{itemize}
\item \textit{Cause}: Identification of the initiator or reason leading to a particular event.
\item \textit{Effect}: Determination of the outcome or consequence resulting from a specific cause.
\end{itemize}

\noindent \textbf{Storytelling}: Prediction of appropriate story endings, with an emphasis on temporal elements.

\subsection{Data Construction}
\textbf{Ordering}
For our ordering dataset, the \textit{facts} problems were derived from actual events extracted from historical timelines on Wikipedia. Specifically, pages such as \url{https://en.wikipedia.org/wiki/Timeline_of_the_18th_century} served as our primary data sources. These timelines cover events ranging from ancient history to the 21st century, offering a rich foundation for our dataset. We explored dedicated pages for each available century, ensuring a diverse collection of events across various epochs.

\noindent \textbf{Frequency}
For the frequency task, three main subtasks are generated based on templates: comparison, computation, and applications. Each template contains placeholders, denoted by \{\}, to represent both events and times. Table~\ref{tab: frequency_templates} outlines some representative templates for each subtask. The construction processes for other subtasks are detailed in the main paper.

\begin{table*}[h]
\centering
\caption{Major templates are used for constructing the frequency subtasks: comparison, computation, and applications. The symbols \{\} serve as placeholders for variable inputs, which can represent both events and times.}
\begin{tabularx}{\textwidth}{lX}
\toprule
\textbf{Category} & \textbf{Template} \\
\midrule
Comparison & Compare the frequency of \{\} and \{\}. \\
\midrule
\multirow{3}{*}{Computation} & If \{\} happens \{\}, how many times will it occur in \{\} years? \\
& \{\} appears \{\}. If it was last seen in \{\}, when will it next appear? \\
& \{\} appears \{\}. If it took place in \{\}, when did it previously occur? \\
\midrule
\multirow{21}{*}{Application} & If a person's job contract has a renewal every \{\} years, and they started working in \{\} and renewed it \{\} times without gaps, until what year is their current contract valid? \\
& A solar eclipse happens at least \{\} times a year. If the first one in \{\} is in \{\}, in which month can we expect the next one? \\
& If a plant blooms every \{\} days and it last bloomed on January 1, on what date will it next bloom? \\
& A comet passes Earth every \{\} years. If its last appearance was in \{\}, when will it next appear? \\
& If a magazine publishes a special edition every \{\} months and the last one was in January, in which month will the next special edition be? \\
& A company holds a general meeting every \{\} quarters. If the last one was in Q1 of a year, which quarter will the next meeting be? \\
& A species of cicada emerges every \{\} years. If they last emerged in \{\}, when will they next emerge? \\
& If a leap year occurs every 4 years and the last one was in \{\}, when is the next leap year? \\
& A festival is celebrated every \{\} years. If it was last celebrated in \{\}, when will it next be celebrated? \\
& If a building undergoes maintenance every \{\} months and the last maintenance was in January, which month will the next maintenance be? \\
\bottomrule
\end{tabularx}
\label{tab: frequency_templates}
\end{table*}

\noindent \textbf{Duration}
For the duration task, five main subtasks are generated based on templates: multi-step comparison, analogy inference, computation, direct comparison, and facts. Each template contains placeholders, denoted by \{\}, to represent both events and times. Table~\ref{tab: duration_templates} outlines some representative templates for each subtask. The construction processes for other subtasks of the dataset are described in the main paper.
\begin{table*}[ht]
\centering
\caption{Major templates for constructing the duration subtasks:multi-step comparison, analogy inference, computation, direct comparison, and facts. The symbols \{\} serve as placeholders for variable inputs, which can represent both events and times.}
\label{tab: duration_templates}
\begin{tabularx}{\textwidth}{lX}
\toprule
\textbf{Type} & \textbf{Template} \\
\midrule
\multirow{10}{*}{Multi-Step Comparison} & \{\} goes on for \{\}. \{\} is a third of \{\}, and \{\} is as long as \{\} and \{\} combined. Which event lasts the longest? \\
& Between \{\} that lasts \{\}, \{\} that is four times longer, and \{\} that's half the total duration of the two, which is the shortest? \\
& \{\} spans \{\}. \{\} is double that, but \{\} is only a third of \{\}. Which has the most extended duration? \\
& If \{\} lasts \{\}, \{\} is twice as long, and \{\} is half of \{\}, which event has the medium duration? \\
& \{\} lasts for \{\}. \{\} is half of \{\}'s duration, and \{\} is triple the combined length of both \{\} and \{\}. Which event has the shortest duration? \\
\midrule
\multirow{15}{*}{Analogy Inference} & During \{\}, the audience had a chance to enjoy a long opera, while \{\} showcased just one act, and \{\} played only an overture. Which event was the shortest? \\
& People could indulge in a seven-course meal during \{\}, a three-course meal in \{\}, but only an appetizer during \{\}. Which event was in the middle in terms of duration? \\
& \{\} felt like watching an epic trilogy, \{\} was more of a feature-length film, and \{\} was just a brief trailer. Which event was probably the longest? \\
& Participants at \{\} went through an entire yoga session, \{\} allowed for a short warm-up, while \{\} was only a few stretches. Which event was the shortest? \\
& During \{\}, attendees could finish a whole board game, in \{\} they played just a few rounds, and in \{\} merely set up the pieces. Which event was likely the longest? \\
\midrule
\multirow{11}{*}{Computation} & The duration of \{\} is \{\}. If \{\} is a quarter shorter than \{\} and \{\} is four times the length of \{\} for \{\}, how long do all the activities last? \\
& For \{\}, \{\} takes \{\}. If \{\} is twice that duration minus 10\% of \{\}, and \{\} is half of the sum of \{\} and \{\}, how long is the whole event? \\
& The total duration of \{\} is four times the time of \{\} which is \{\}. If \{\} is half of \{\} minus 5\% of \{\} and \{\} is twice \{\} plus 15\% of \{\}, how long do the \{\} and \{\} together take? \\
& In \{\}, \{\} is twice as long as \{\} which takes \{\}. If \{\} is the difference between \{\} and \{\}, how long in total? \\
& For \{\}, \{\} lasts for \{\}. If \{\} is double that duration minus 15\% of \{\} and \{\} is the sum of \{\} and \{\} divided by 2, what's the entire duration? \\
\midrule
\multirow{2}{*}{Direct Comparison} & Which event lasted longer: \{\} or \{\}? \\
& Which event lasted the longest: \{\}, \{\}, or \{\}? \\
\midrule
Facts & How long did \{\} last? \\
\bottomrule
\end{tabularx}
\end{table*}

\noindent \textbf{Typical Time}
For the typical time task, we crafted pairs of time-related events to test the model's proficiency in determining “Which statement is more typical in terms of time?” For instance, when presented with statements such as “People often have dinner in the early to late evening” and “People often have dinner in the mid to late afternoon”, the model is prompted to recognize which one is more aligned with a conventional behavior. Similarly, it might evaluate statements like “Bars are typically busiest on Friday and Saturday nights” in comparison to “Bars are typically busiest on Sunday and Monday nights”. Through these examples, we aim to assess the model's aptitude in discerning standard temporal practices.

\noindent \textbf{Ambiguity Resolution}
For the ambiguity resolution task, we introduced templates to test the model's proficiency in resolving temporal ambiguities. Additionally, we manually gathered both common and uncommon temporal expressions that might perplex individuals and the model alike, such as “for a coon's age”, “when pigs fly”, and “in the nick of time”. Table~\ref{tab: ambiguity_templates} presents representative templates for each subtask. Each template contains placeholders, denoted by \{\}, to represent both events and times.

\begin{table*}[h]
\centering
\caption{Major templates used for constructing the ambiguity resolution dataset. The symbols \{\} serve as placeholders for variable inputs, which can represent both events and times.}
\label{tab: ambiguity_templates}
\begin{tabularx}{\textwidth}{lX}
\midrule
\textbf{Type} & \textbf{Template} \\
\midrule
\multirow{6}{*}{Short-term Shift} & Your plane is supposed to depart at \{\}. If it's preponed by \{\}, when is the revised departure? \\
& The meal was promised to be on the table at \{\}. If it's going to be \{\} postponed, when can you expect to dine? \\
& You have an exciting date at \{\}. If you're lagging by \{\}, when will you probably meet your date? \\
\midrule
\multirow{11}{*}{Mid-term Shift} & The match initially set for \{\} has now been advanced by \{\}. Which day is it on now? \\
& Your usual spa day on \{\} of every week has been postponed \{\}. When will it be next week? \\
& The weekly town hall usually on \{\} is delayed by \{\}. When will it happen? \\
& The town carnival usually during the \{\} week of \{\} will now be \{\}. About which date is it now? \\
& The music fest during the \{\} week of \{\} will be held \{\}. Around which date will it likely be? \\
& The product launch in the \{\} week of \{\} has been shifted \{\}. Around when will it likely be? \\
\midrule
\multirow{4}{*}{Long-term Shift} & The star, predicted to explode in \{\}, has its explosion postponed by \{\} years. When is the new prediction for the explosion? \\
& The dynasty which fell in \{\} had risen to power roughly \{\} years earlier. When was its establishment? \\
\midrule
Calendar Shift & If the date is \{\}/\{\}/\{\} in the \{\}, what is the date in the \{\}? \\
\midrule
\multirow{3}{*}{Interpretation} & You receive a memo with the timestamp \{\}. When should you be prepared? \\
& A festival is being organized \{\}. When would that be? \\
& A note suggests meeting \{\}. When is this suggesting? \\
\bottomrule
\end{tabularx}
\end{table*}

\noindent \textbf{Arithmetic}
We mainly adopted a programmatic generation approach, grounded in meticulously designed templates that focus on specific temporal calculations. These templates encompass a variety of temporal arithmetic tasks, ranging from basic time adjustments to more complex calculations like week identifications and real-world applications. Table~\ref{tab: arithmetic_templates} shows the major templates we use for constructing the arithmetic datasets. The variable values, denoted by \{\}, are randomly generated by programs. Through these templates, we can generate diverse questions that test a model's proficiency in handling different temporal arithmetic scenarios.

\begin{table*}[h]
\centering
\caption{Major templates used for constructing the arithmetic dataset. The symbols \{\} serve as placeholders for variable inputs, which are randomly generated by programs.}
\label{tab: arithmetic_templates}
\begin{tabularx}{\textwidth}{lX}
\toprule
\textbf{Category} & \textbf{Template} \\
\midrule
24-hour Adjustment & What is \{\}:\{\} +/- \{\}:\{\}? \\
\midrule
12-hour Adjustment & What is \{\}:\{\} AM/PM +/- \{\}:\{\}? \\
\midrule
\multirow{2}{*}{Year Shift} & Which year comes \{\} years after \{\}? \\
& Which year was \{\} years before \{\}? \\
\midrule
\multirow{2}{*}{Month Shift} & Which month comes \{\} months after \{\}? \\
& Which month was \{\} months before \{\}? \\
\hline
\multirow{6}{*}{Date Computation} & What will be the time \{\} years and \{\} months after month \{\}? \\
& If you add/subtract \{\} days to the date \{\}, what will be the new date? \\
& If you add/subtract \{\} months and \{\} days to the date \{\}, what will be the new date? \\
& If you add/subtract \{\} weeks and \{\} days to the date \{\}, what will be the new date? \\
\midrule
Week Identification & In which week of year \{\} does the date \{\} occur? \\
\midrule
Time Zone Conversion & If it’s \{\} in the source zone, what's the date and time in target zone? \\
\midrule
\multirow{6}{*}{Time Computation} & Convert \{\} days into minutes. \\
& Convert \{\} minutes into hours. \\
& Convert \{\} days into hours. \\
& Convert \{\} seconds into hours. \\
& Add \{\} minutes \{\} seconds and \{\} minutes \{\} seconds. \\
& Subtract \{\} minutes \{\} seconds from \{\} hours \{\} minutes. \\
\midrule
\multirow{13}{*}{Application} & If a person takes a leave of \{\} days starting from start\_date, on which day may the leave end? \\
& If a person was \{\} years \{\} month(s) old when he joined school and now he is \{\} years \{\} month(s) old, for how long has he been in school? \\
& If a person is advised to take medicine every \{\} minutes, how many times will she take the medicine in a day? \\
& If a person starts doing homework at \{\} and finishes at \{\} PM, how many hours did he spend on homework? \\
& If a flight takes off at \{\} and the duration of the flight is \{\} hours, at what time will it land? \\
& If a person walks at a speed of \{\} km/hr and after every km, she takes a rest for \{\} minutes, how many minutes will it take her to cover \{\} km? \\
& How long will it take to travel a distance of \{\} kilometers in minutes? \\
\midrule
\end{tabularx}
\end{table*}

\noindent \textbf{Relation}
To derive temporal relation questions from the TempEval-3 Silver dataset, we iterated through each temporal link (\textit{tlink}) to extract the relationship type (\textit{relType}) and relevant event and time IDs. For each \textit{tlink}, the associated \textit{eventInstanceID} provided the \textit{eventID}, either directly or via the \textit{MAKEINSTANCE} tag. We then identified the sentence containing this event as its contextual background. Using the gathered data, we crafted questions such as “What is the relationship between the event ‘${event_1}$' and the event ‘${event_2}$'?” or analogous questions pertaining to event-time relationships. The context, encompassing both events, was attached to the resulting question to ensure clarity.

\noindent \textbf{Temporal NLI}
To construct our temporal NLI dataset, we adopted a keyword-based filtering approach from SNLI and MNLI datasets. Recognizing that NLI tasks can often hinge on nuanced temporal cues, we curated a comprehensive set of temporal keywords, as shown in Table~\ref{tab: nli_temporal_keywords}. This selection was designed to capture a broader range of temporal relationships and nuances. Instances containing at least one term from this extended list were considered to possess temporal elements and were thus included for further analysis.

\begin{table*}[h]
\centering
\caption{Keywords used for filtering SNLI and MNLI datasets that contain temporal aspects.}
\label{tab: nli_temporal_keywords}
\begin{tabularx}{\linewidth}{lX}
\toprule
\textbf{Category} & \textbf{Keywords} \\
\midrule
\multirow{3}{*}{Explicit References} & today, tomorrow, yesterday, now, soon, later, before, after, day, week, month, year, hour, minute, second, morning, evening, night, noon, midnight, anniversary \\
\midrule
\multirow{2}{*}{Days of the Week} & Monday, Tuesday, Wednesday, Thursday, Friday, Saturday, Sunday \\
\midrule
\multirow{2}{*}{Months} & January, February, March, April, May, June, July, August, September, October, November, December \\
\midrule
Seasons & spring, summer, fall, autumn, winter \\
\midrule
Periods and Eras & decade, century, millennium, epoch, era \\
\midrule
\multirow{2}{*}{General Terms} & annual, biannual, quarterly, hourly, daily, weekly, quarter, monthly, fortnight, biweekly, bimonthly, semester, trimester \\
\midrule
\multirow{2}{*}{Relative Terms} & past, future, current, upcoming, recent, lately, ago, in advance, later, previous, next, moment, time, when, while, duration, period, early, earlier \\
\midrule
\multirow{2}{*}{Implicit Temporal Actions} & wait, postpone, delay, reschedule, expire, due, schedule, begin, start, end, finish, commence, conclude, last, extend \\
\midrule
\multirow{2}{*}{Temporal Transitions and Connectors} & until, by the time, as soon as, whenever, since, during, whilst \\
\midrule
\multirow{3}{*}{Other Temporal Entities} & sunset, sunrise, dusk, dawn, midday, eve, annually, eventually, seldom, often, always, never, sometimes, usually, frequently, occasionally, rarely, just, once, still \\
\midrule
\end{tabularx}
\end{table*}

\noindent \textbf{Causality}
Inspired directly by the style of the COPA dataset, our goal was to capture the intricate weave of cause-and-effect relationships shaped by temporal elements. To this end, we prioritized the inclusion of diverse temporal factors in our dataset, encompassing aspects like seasons, specific times on clocks, special occasions, as well as both long-term and short-term causes and impacts. By meticulously crafting problems with these considerations, we have crafted a rich collection that illuminates the nuanced interplay between time and causality.

\noindent \textbf{Storytelling}
To identify stories with temporal nuances from the ROCStories and SCT datasets, we employed a keyword-based filtering approach. The choice of our keyword set, as shown in Table~\ref{tab: story_temporal_keywords}, was shaped by the distinctive nature of the datasets and the contexts they encompass. In ROCStories, for instance, storytelling often employs varied and colloquial temporal expressions, necessitating a specific focus in our keyword selection. Stories containing at least one term from the list were considered to have temporal aspects and were subsequently selected for further processing.

\begin{table*}[ht]
    \centering
    \caption{Keywords used for filtering ROCStories and SCT datasets that contain temporal aspects.}
    \begin{tabular}{ll}
    \toprule
    \textbf{Category} & \textbf{Keywords} \\
    \midrule
    Time References & \makecell[l]{before, after, recently, now, then, \\ earlier, later, today, tonight, \\ yesterday, tomorrow} \\
    \midrule
    Temporal Intervals & \makecell[l]{soon, nowadays, currently, presently, \\ eventually, ultimately, suddenly, \\ immediately, momentarily, previously, formerly} \\
    \midrule
    Recurring Time Periods & \makecell[l]{periodically, seasonally, daily, weekly, \\ monthly, annually, biennially} \\
    \midrule
    Fixed Time Periods & \makecell[l]{century, decade, millennium, year, \\ minute, hour, day, week, month} \\
    \midrule
    Parts of the Day & \makecell[l]{morning, noon, evening, night} \\
    \midrule
    Duration \& Frequency & \makecell[l]{duration, instant, temporarily, intermittently, \\ frequently, always, never, sometimes, \\ often, rarely, usually} \\
    \midrule
    Starting Actions & \makecell[l]{begin/begins/began, start/starts/started, \\ commence/commences/commenced} \\
    \midrule
    Ending Actions & \makecell[l]{end/ends/ended, finish/finishes/finished, \\ cease/ceases/ceased, expire/expires/expired, \\ elapse/elapses/elapsed} \\
    \midrule
    Continuing \& Delaying & \makecell[l]{last/lasts/lasted, continue/continues/continued, \\ resume/resumes/resumed, linger/lingers/lingered, \\ postpone/postpones/postponed, \\ procrastinate/procrastinates/procrastinated} \\
    \bottomrule
    \end{tabular}
    \label{tab: story_temporal_keywords}
\end{table*}

\subsection{Example Questions}
\begin{figure*}[ht]
\centering
\includegraphics[width=\textwidth]{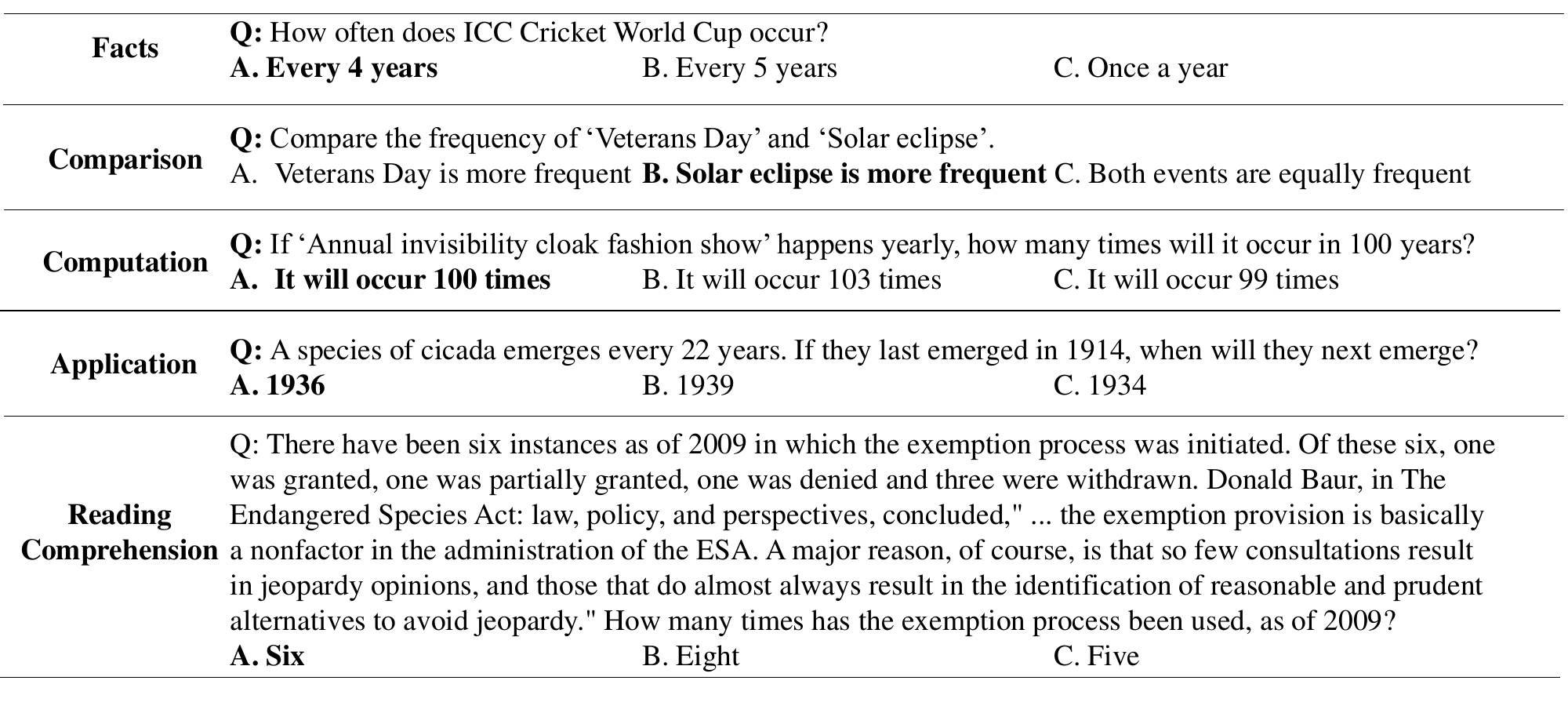}  
\caption{Example questions on the frequency task.}
\label{fig: more_frequency}
\end{figure*}

\begin{figure*}[ht]
\centering
\includegraphics[width=\textwidth]{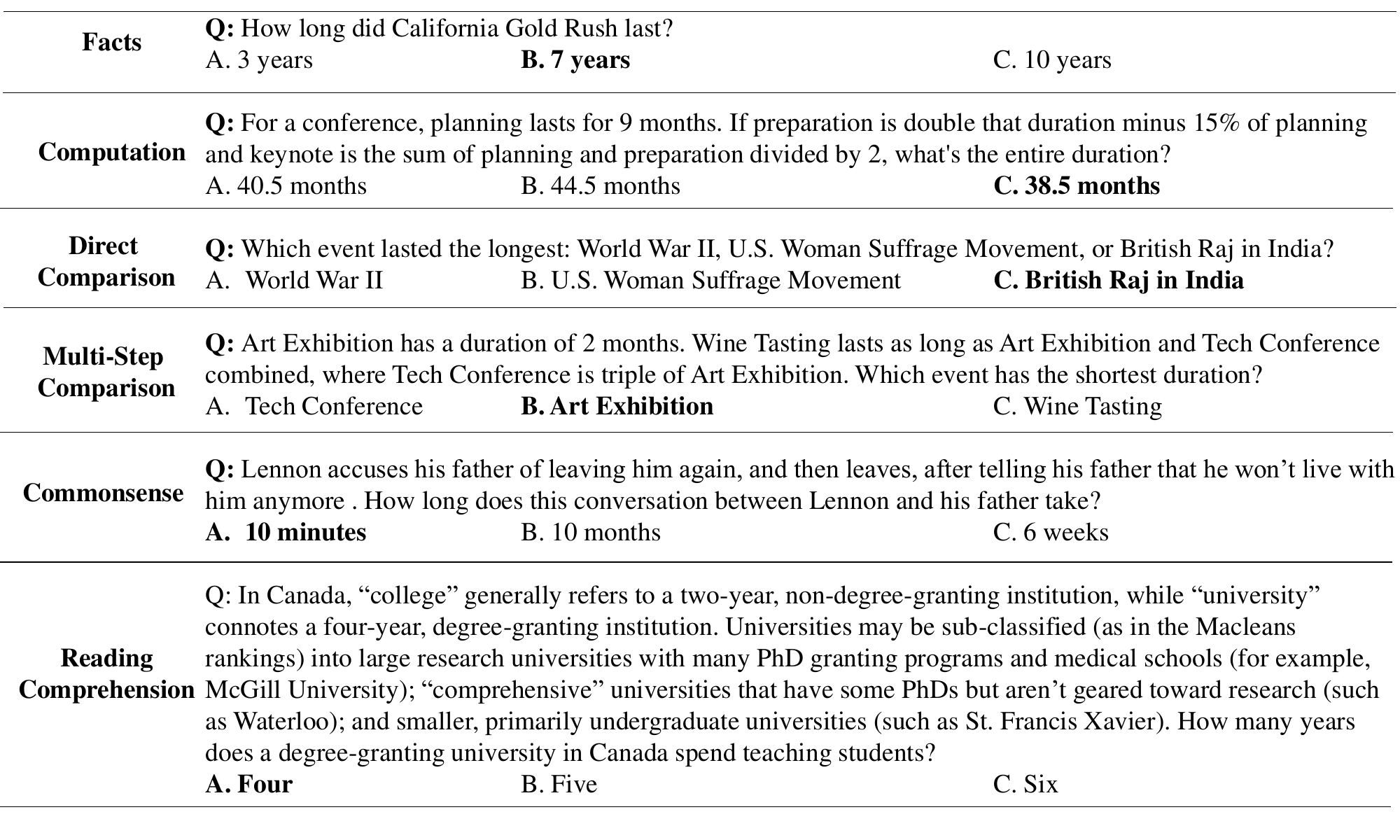}  
\caption{Example questions on the duration task.}
\label{fig: more_duration}
\end{figure*}

\begin{figure*}[ht]
\centering
\includegraphics[width=\textwidth]{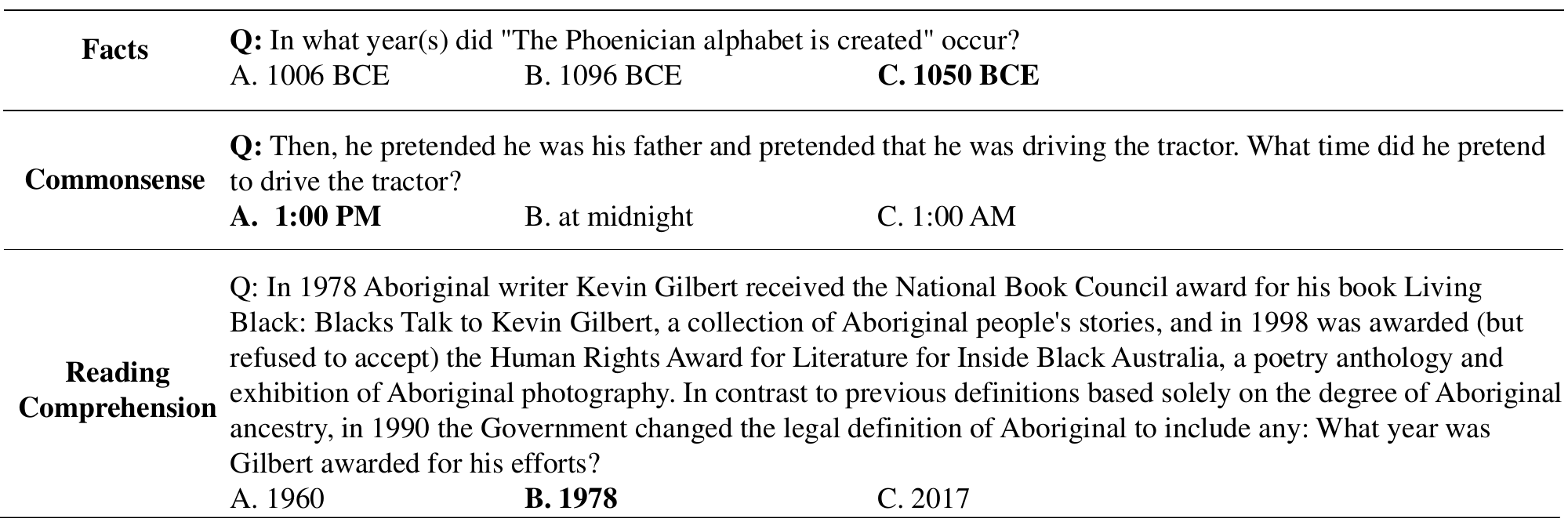}  
\caption{Example questions on the typical time task.}
\label{fig: more_typical_time}
\end{figure*}

\begin{figure*}[ht]
\centering
\includegraphics[width=\textwidth]{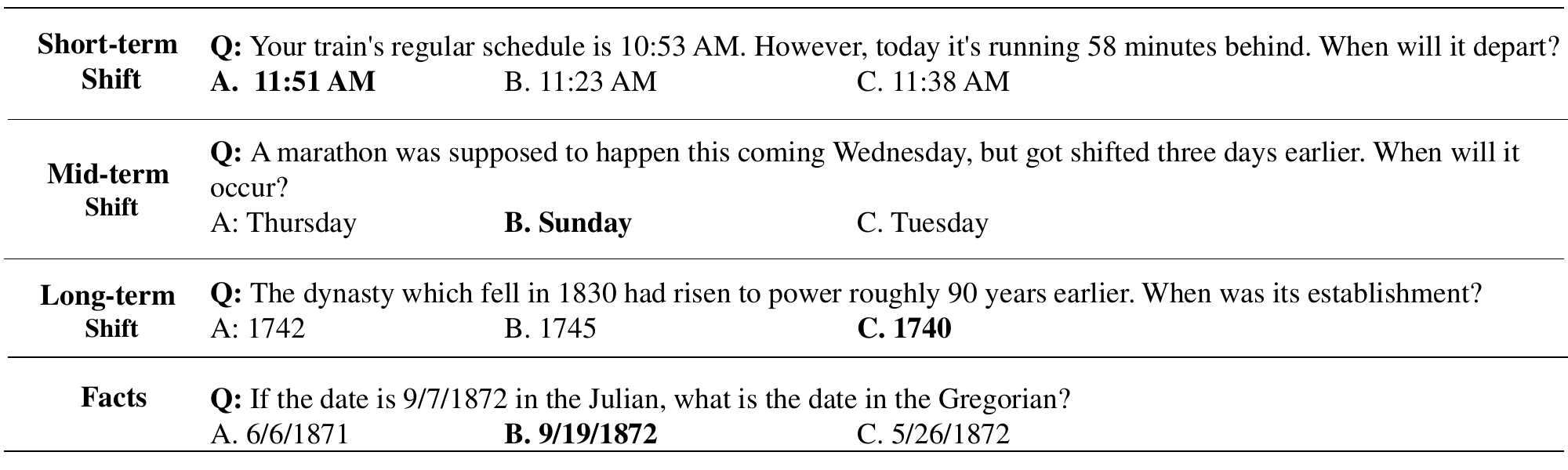}  
\caption{Example questions on the ambiguity resolution task.}
\label{fig: more_ambiguity_resolution}
\end{figure*}

\begin{figure*}[ht]
\centering
\includegraphics[width=\textwidth]{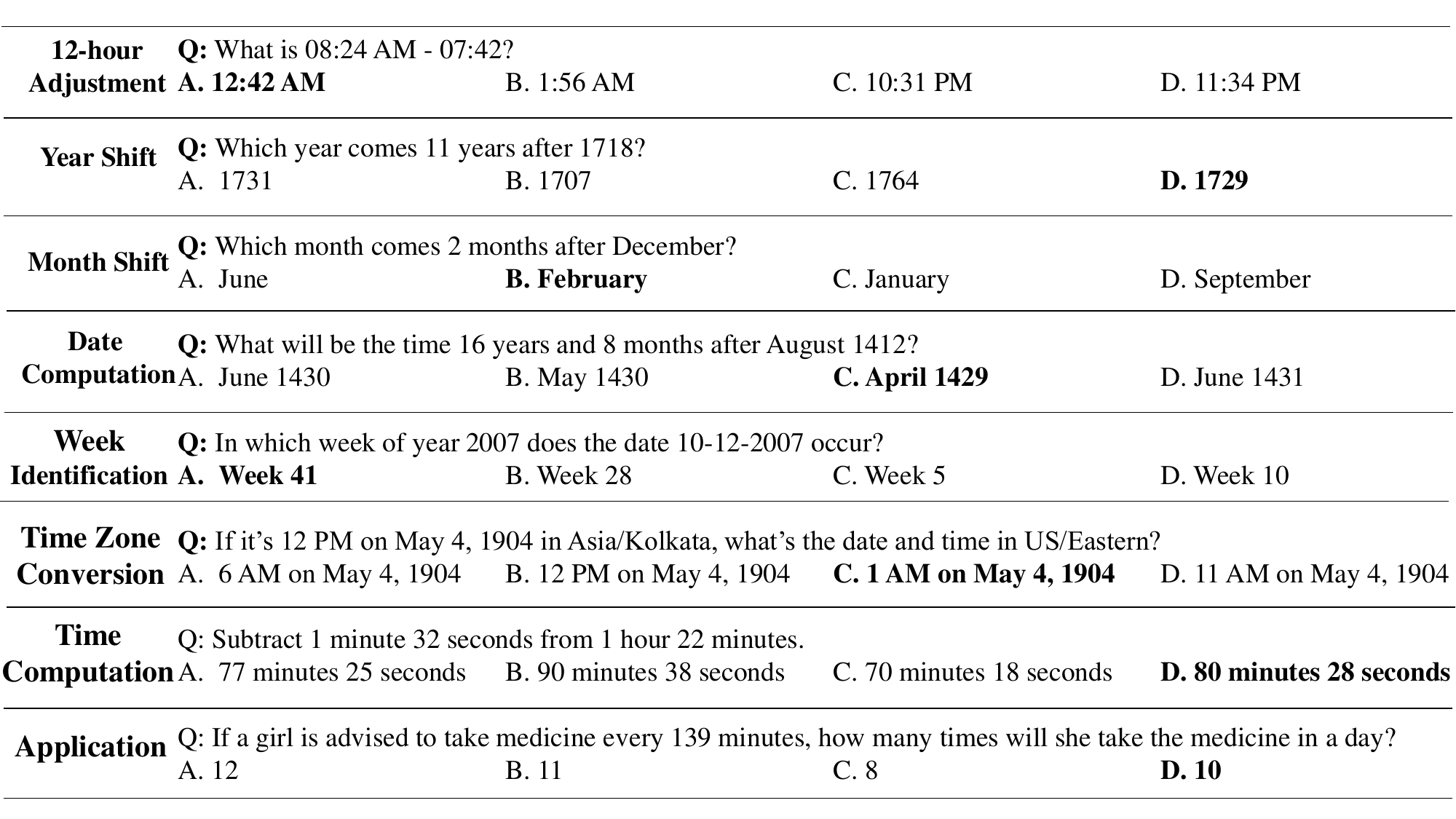}  
\caption{Example questions on the arithmetic task.}
\label{fig: more_arithmetic}
\end{figure*}

\begin{figure*}[ht]
\centering
\includegraphics[width=\textwidth]{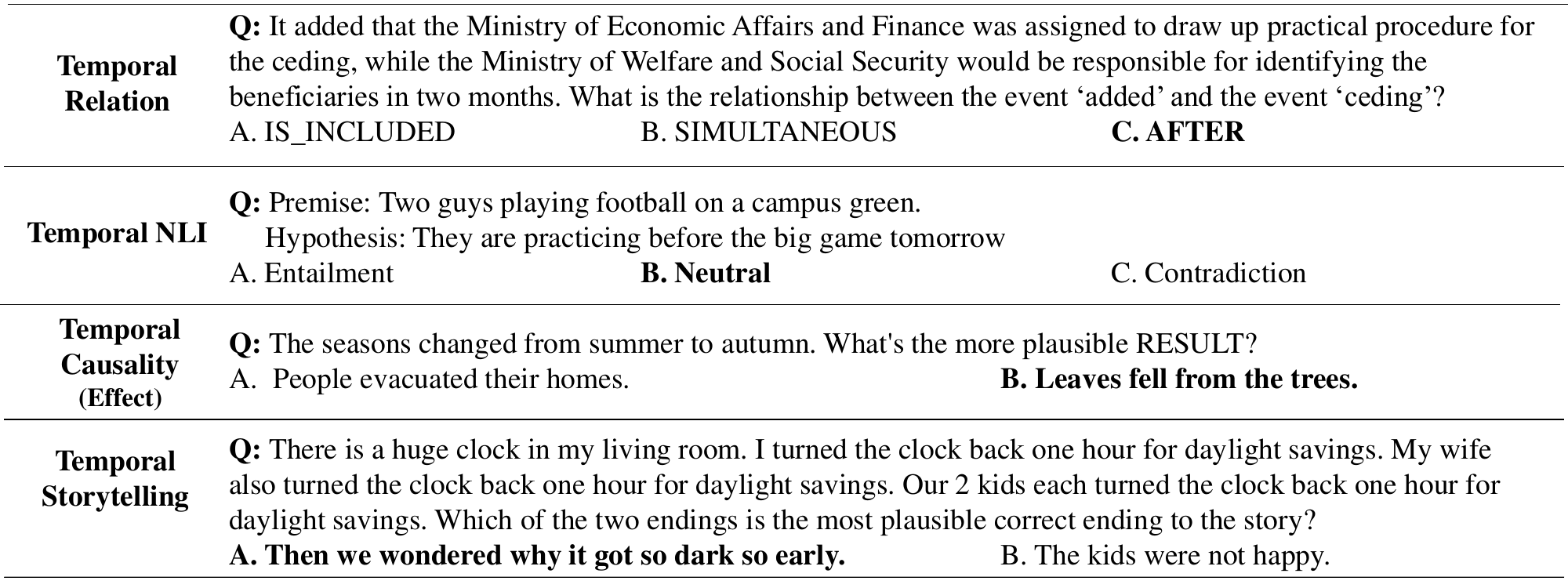}  
\caption{Example questions on advanced temporal reasoning tasks, including relation, temporal NLI, causality, and storytelling.}
\label{fig: more_misc}
\end{figure*}

For additional examples of various tasks, refer to the following figures: Figure~\ref{fig: more_ordering} for the ordering task, Figure~\ref{fig: more_frequency} for the frequency task, Figure~\ref{fig: more_duration} for the duration task, Figure~\ref{fig: more_typical_time} for the typical time task, Figure~\ref{fig: more_ambiguity_resolution} for the ambiguity resolution task, and Figure~\ref{fig: more_arithmetic} for the arithmetic task. The advanced temporal understanding group, comprising relation, temporal NLI, causality, and storytelling tasks, which have relatively fewer subtasks, are collectively presented in Figure~\ref{fig: more_misc}. The correct choices are bolded.

\begin{figure*}[ht]
\centering
\includegraphics[width=\textwidth]{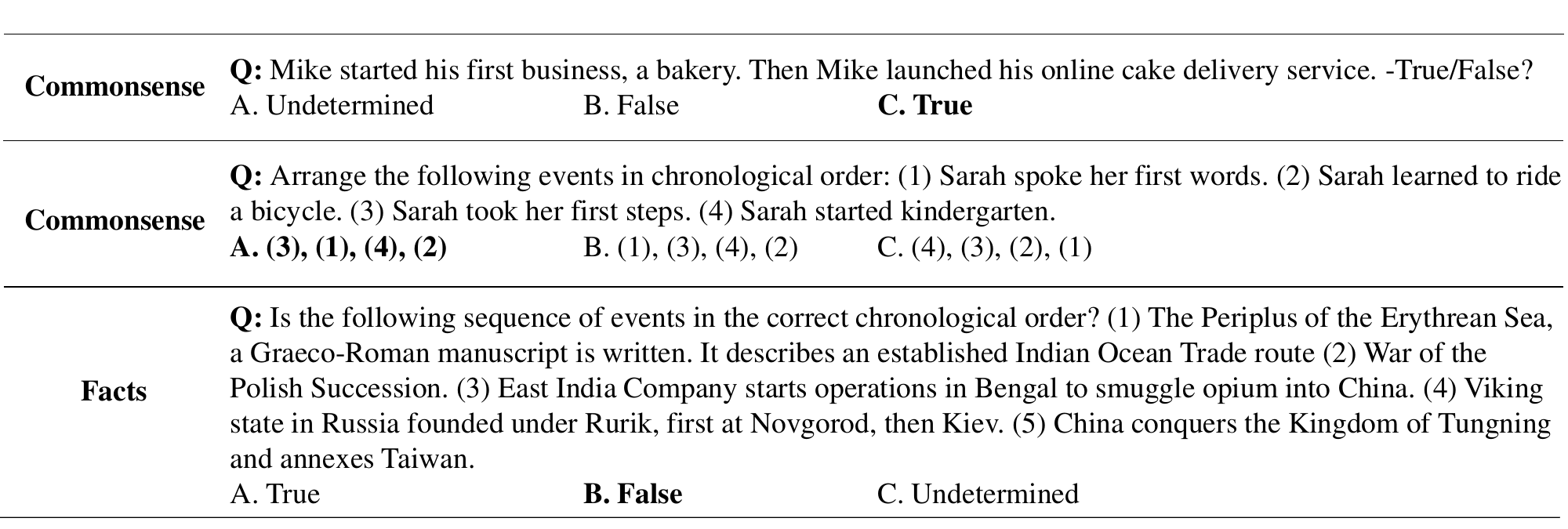}  
\caption{Example questions on the temporal ordering task.}
\label{fig: more_ordering}
\end{figure*}

\subsection{Comparison of Source vs. Curated Datasets}
We provide several representative examples sourced from existing datasets, allowing for a comparison between the original sources and our curated datasets. Specifically, Table~\ref{tab: MCTACO_comp1} and Table~\ref{tab: MCTACO_comp2} demonstrate the transformation of original Yes/No binary questions from the MCTACO dataset into our frequency and ordering tasks in MCQ formats, respectively. Meanwhile, Table~\ref{tab: SQuAD_comp} shows the transformation of original short-answer questions from the SQuAD dataset into our duration task. Our benchmark combines the strengths of existing benchmarks with extensive manual effort, including the addition of distracting or confusing options, the filtering out of irrelevant questions for quality control, and the reformulation of problems, thereby setting a new standard for assessing temporal reasoning in LLMs.

\begin{table*}[h]
\centering
\caption{{Comparison of source (MCTACO) and curated question in TRAM for the Frequency task.}}
\begin{tabular}{p{0.45\linewidth} p{0.45\linewidth}}
\hline
\textbf{Source Dataset (MCTACO)} & \textbf{Curated Dataset (TRAM)} \\
\hline
\textbf{Question:} Allan crouched over his desk once more, pen in hand and mind blank. How often does Allan crouch over his desk? & \textbf{Question:} Allan crouched over his desk once more, pen in hand and mind blank. How often does Allan crouch over his desk? \\
\hline
\textbf{Options/Answers:}
\begin{itemize}
    \item Once a second - No
    \item Once two years ago - No
    \item Every day - Yes
    \item Several times per second - No
    \item Daily - Yes
\end{itemize} & \textbf{Options:}
\begin{itemize}
    \item (A) Every day
    \item (B) Several times per second
    \item (C) Once a second
\end{itemize}
\textbf{Answer:}
\begin{itemize}
    \item (A) Every day
\end{itemize} \\
\hline
\textbf{Commentary:} Binary Yes/No format, simple frequency assessment. & \textbf{Commentary:} Transition to an MCQ format enriches the question's complexity by offering closely related alternatives. \\
\hline
\end{tabular}
\label{tab: MCTACO_comp1}
\end{table*}

\begin{table*}[h]
\centering
\caption{{Comparison of source (MCTACO) and curated question in TRAM for the Ordering task.}}
\begin{tabular}{p{0.45\linewidth} p{0.45\linewidth}}
\hline
\textbf{Source Dataset (MCTAC0)} & \textbf{Curated Dataset (TRAM)} \\
\hline
\textbf{Question:} Church is brought back to life, but is an evil shell of himself. What did Church do next? & \textbf{Question:} Church is brought back to life, but is an evil shell of himself. What did Church do next? Is ”took a nap” possible? \\
\hline
\textbf{Options/Answers:}
\begin{itemize}
    \item ”took a nap” - No
\end{itemize} & \textbf{Options:}
\begin{itemize}
    \item (A) Undetermined
    \item (B) TRUE
    \item (C) FALSE
\end{itemize} 
\textbf{Answer:}
\begin{itemize}
    \item (B) Two months
\end{itemize} \\
\hline
\textbf{Commentary:} Binary Yes/No format, simple ordering assessment. & \textbf{Commentary:} Transition to an MCQ format introduces additional ambiguity and uncertainty into the question.\\
\hline
\end{tabular}
\label{tab: MCTACO_comp2}
\end{table*}

\begin{table*}[h]
\centering
\caption{{Comparison of source (SQuAD) and curated question in TRAM for the Duration task.}}
\begin{tabular}{p{0.45\linewidth} p{0.45\linewidth}}
\hline
\textbf{Source Dataset (SQuAD)} & \textbf{Curated Dataset (TRAM)} \\
\hline
\textbf{Question:} It was not until January 1518 that friends of Luther translated the 95 Theses ... Within two weeks, copies of the theses had spread throughout Germany; within two months, they had spread throughout Europe. How long did it take for the Theses to spread through Europe? & \textbf{Question:} It was not until January 1518 that friends of Luther translated the 95 Theses ... Within two weeks, copies of the theses had spread throughout Germany; within two months, they had spread throughout Europe. How long did it take for the Theses to spread through Europe? \\
\hline
\textbf{Options/Answers:}
\begin{itemize}
    \item Short answer: Two months
\end{itemize} & \textbf{Options:}
\begin{itemize}
    \item (A) 45 days
    \item (B) Two months 
    \item (C) 2 days
\end{itemize} 
\textbf{Answer:}
\begin{itemize}
    \item (B) Two months
\end{itemize} \\
\hline
\textbf{Commentary:} Short-answer format, simple duration assessment. & \textbf{Commentary:} Transition to an MCQ format introduces additional numerical ambiguity in problems involving multiple numbers. \\
\hline
\end{tabular}
\label{tab: SQuAD_comp}
\end{table*}

\subsection{Subtask Distributions}
\begin{figure*}[ht]
\centering
\includegraphics[width=\textwidth]{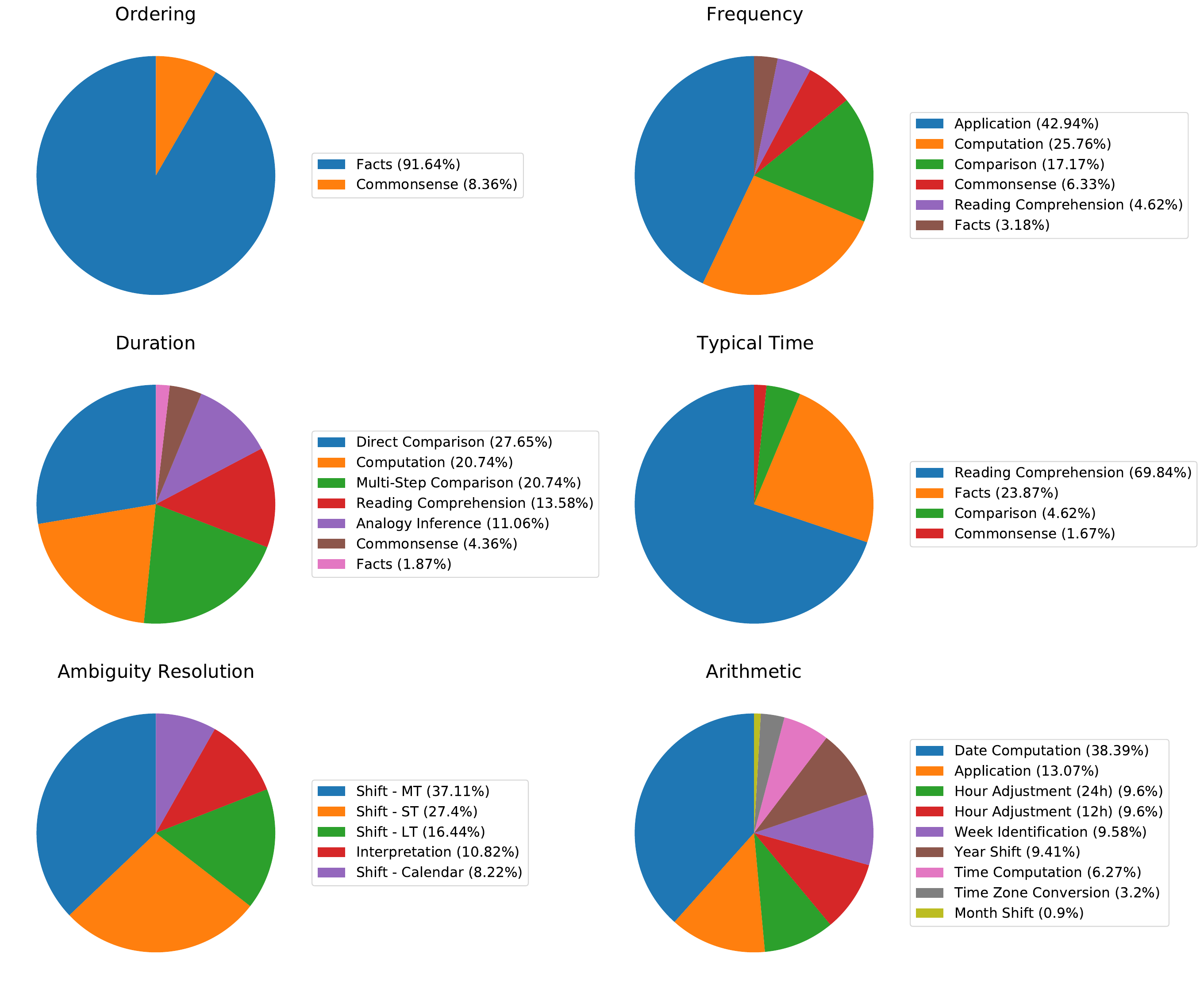}  
\caption{Distribution of subtasks for each distinct temporal reasoning task.}
\label{fig: subtask_distribution}
\end{figure*}

As shown in Table~\ref{tab: dataset}, if \textit{Problem Types} count exceeds 1, then we consider it a task involving multiple subtasks. Figure~\ref{fig: subtask_distribution} illustrates the distribution of subtasks for each temporal reasoning task. In the case of causality, two problem types are evenly distributed, each accounting for 50\%.

\section{Prompts}~\label{prompts} 
We utilize both SP and CoT in our experiments with LLMs. For SP, questions are presented directly without the need for additional steps in the prompt. Consider the following example from the storytelling dataset:

“When I was a boy, my parents used to take my brother and me to the park. We would play, have lunch, and just walk around. One day, when all the picnic benches at the park were occupied, we had one. Two police officers approached and asked if they could join us. Which of the two endings is the most plausible correct ending to the story?

(A) They were there to take my brother and me to the police station. \\
(B) They let us operate the police car lights and siren.”

For zero-shot SP, the model is simply prompted with the question: “Given the story ‘When I was a boy ... they could join us.' Which of the two endings is the most plausible correct ending to the story? (A) They were... or (B) They let us... The answer (A or B) is: \{ \}.” For few-shot SP, exemplar answers (A or B) are provided alongside the questions.

The overall SP procedure across all tasks can be summarized in three steps: (1) \textit{Context Provision (if any):} Provide any necessary background information or context that may aid the model in understanding the scenario presented in the question. (2) \textit{Direct Questioning:} Pose the question directly to the model without any intermediary steps or additional guidance. (3) \textit{Answer Solicitation:} Request the model to choose and provide the most appropriate answer based on the information given.

In contrast, for CoT, zero-shot learning takes inspiration from \citep{kojima2022large} by instructing the model to “Answer the question step by step”. For few-shot CoT, we manually craft the step-by-step process for 5-shot exemplars in the development set. 
The procedure to approach this problem is as follows:

\begin{enumerate}
\item[(1)] \textit{Read the Story Carefully:} Understand the main theme, setting, and characters introduced in the story. The dominant theme appears to be a nostalgic recollection of a family day out at a park.
\item[(2)] \textit{Identify Key Elements from the Story:} The protagonist recalls a childhood memory. The primary setting is a park. The mood is both casual and reminiscent. Despite the park being crowded, they have a picnic spot. Subsequently, two police officers approach the family.
\item[(3)] \textit{Evaluate Each Proposed Ending:} For the first ending, a sudden and unexpected twist is introduced that deviates from the story's initial light-hearted narrative. This ending lacks context about why they'd be taken to the police station. The second ending maintains the story's casual and friendly tone, presenting a scenario where the police officers engage positively with the family.
\item[(4)] \textit{Comparison of the Two Endings:} Both endings involve the police officers, but the first one introduces a jarring twist without adequate prior context. The second ending aligns more consistently with the story's overarching mood and theme.
\item[(5)] \textit{Conclusion:} Given the story's tone, setting, and characters, the second ending appears more plausible and contextually appropriate.
\end{enumerate}

After defining the step-by-step procedure, we employ it to steer the model's thought process. This structured methodology better prepares the model to reason through the question and formulate a well-considered answer, thereby providing a distinct advantage over the SP method. We structure our prompt as follows: “Begin by reading the story carefully, ensuring you fully understand its main theme, setting, and the characters. \{Immediate analysis\}. Subsequently, identify the key elements of the story. \{Immediate analysis\}. Assess each proposed ending within the context of the narrative. \{Immediate analysis\}. Compare the two endings, highlighting any thematic or tonal discrepancies. \{Immediate analysis\}. Conclude by determining which ending appears more plausible, offering a rationale for this selection \{Immediate analysis\}.”

In general, the CoT procedure across all temporal reasoning tasks is as follows: (1) \textit{Understanding Context:} Begin by reading the provided data, statement, or story attentively. Understand the overarching theme, objectives, or the problem's primary ask. (2) \textit{Key Elements Extraction:} Identify and highlight crucial elements, specifics, or characters. This could mean different things for different tasks - key events in a story, terms in a mathematical problem, or clauses in a statement. (3) \textit{Evaluation:} Assess the core objective of the problem in its context. This could be understanding the chronology for ordering, assessing frequency, gauging durations, or even understanding the logical or causal flow in more complex problems. (4) \textit{Analysis and Comparison:} If there are multiple options or scenarios presented, conduct a deep analysis. Compare, contrast, and evaluate based on the preceding steps. (5) \textit{Reasoned Conclusion:} Conclude with a structured answer or resolution to the problem, ensuring that the decision-making process aligns with the evidence or data presented. In practice, the procedure varies for each task to account for the diverse nature of temporal reasoning tasks.

\section{Human Assessment}~\label{human_expertise}
In this section, we provide additional details on human participation in our benchmark, including the selection process for experts, verification of their capabilities, and a performance comparison with non-specialists.

\noindent \textbf{Selection of Expert Annotators} Our selection criteria for expert annotators emphasized a balanced proficiency in both temporal reasoning and quantitative analysis. We included professionals with advanced degrees (M.S. or Ph.D.) in disciplines that offer distinct perspectives on our tasks. This included cognitive science and psychology for qualitative understanding of human temporal cognition, crucial for interpreting more subjective aspects of the tasks. We also involved experts in statistics, mathematics, and computer science to address the quantitative complexities inherent in many of our benchmark tasks. This diverse expertise ensured a comprehensive evaluation of the problems within the TRAM dataset from both qualitative and quantitative angles.

\noindent \textbf{Expertise Verification Process} To ensure the high caliber of our expert panel, we implemented a robust screening process. This involved a thorough validation of their educational qualifications and a careful review of their professional and research experience, particularly focusing on time perception and quantitative problem-solving. Additionally, we administered a preliminary assessment composed of one random problem from each subtask, totaling 37 problems. The passing criterion for this assessment was set at an average accuracy rate of more than 92\%, allowing a maximum of three incorrect responses. This stringent benchmark was established to guarantee the experts' capability in accurately addressing the complex problems in TRAM.

\noindent \textbf{Comparison with Unspecialized Individuals} In addition to expert assessments, we conducted a comparative analysis with human non-specialists to provide a broader perspective on human performance. These non-specialists, sourced from Amazon Mechanical Turk, consisted of individuals without specialized training in temporal reasoning or related fields. They were tasked with responding to the same set of 1,900 questions as the experts. This group achieved an overall accuracy rate of 63.5\% across all tasks. This comparison not only underlines the proficiency of our expert panel but also offers insights into the general human ability to tackle TeR challenges, providing a baseline for non-expert performance in this area.

\newpage
\section{Error Types}~\label{error_types}
In this section, we delve into each specific error that LLMs commonly encounter in temporal reasoning tasks, as illustrated in Figure~\ref{fig: errors}.

\noindent \textbf{Foundational Temporal Understanding Tasks} In foundational temporal understanding, LLMs encounter several distinct challenges. Firstly, \textit{Assumption Bias} is evident when models over-rely on patterns from their training, often neglecting cultural or individual variations. Next, \textit{Temporal Descriptor Misinterpretation} occurs when models misinterpret terms, such as perceiving “often” as a daily event instead of a possible weekly occurrence. \textit{Event Ambiguity} presents another challenge, where events can be described in ways that allow for multiple interpretations, requiring models to select the most suitable one based on context. Lastly, \textit{Contextual Misjudgment} is when models either miss or misinterpret explicit temporal clues, leading to errors in their reasoning.

\noindent \textbf{Temporal Interpretation and Computation Tasks} In computational and interpretable temporal reasoning, LLMs encounter various challenges. Firstly, \textit{Calculation Slips} highlight instances where models often make calculation mistakes like inappropriate handling of time carries. Following this, \textit{Descriptor Confusion} arises when models misalign qualitative terms such as “seldom” or “frequently” with their quantitative meanings. \textit{Resolution Misalignment} represents the struggle models face with vague time references, such as deciphering the exact duration from terms like “in a while”. Lastly, \textit{Temporal Notation Misinterpretation} occurs when models confuse time formats, for example, mixing up AM with PM or not differentiating between 24-hour and 12-hour representations.

\noindent \textbf{Advanced Temporal and Conceptual Understanding Tasks} In advanced temporal reasoning tasks, LLMs frequently encounter certain pitfalls. Among the most prevalent is \textit{Implicit Oversights}, where models overlook subtle but crucial temporal indications, resulting in inaccurate conclusions. Also, they may face \textit{Relation Simplification}, wherein complex temporal interplays between events are either misunderstood or overly simplified. LLMs might also fall into the trap of \textit{Narrative Bias}, where they overly depend on familiar story patterns, prioritizing recognized sequences over fresh interpretations. Lastly, \textit{Overgeneralization} becomes evident when models incorrectly apply broad temporal conventions to specific situations, leading to misunderstandings when scenarios diverge from the norm.

\end{document}